\documentclass[journal]{IEEEtran}
\usepackage{amsmath,amsfonts}
\usepackage{algorithmic}
\usepackage{algorithm}
\usepackage{array}
\usepackage[caption=false,font=normalsize,labelfont=sf,textfont=sf]{subfig}
\usepackage{textcomp}
\usepackage{stfloats}
\usepackage{url}
\usepackage{verbatim}
\usepackage{graphicx}
\usepackage{cite}
\usepackage[table]{xcolor}
\usepackage{comment}

\usepackage{balance}
\usepackage{colortbl}
\usepackage{orcidlink}
\usepackage{url}
\usepackage{hyperref}
\usepackage{makecell}
\usepackage{pifont}
\usepackage[table]{xcolor}
\usepackage[utf8]{inputenc}
\usepackage{pgfplots}
\usepgfplotslibrary{colormaps} 
\usetikzlibrary{pgfplots.colormaps}
\usetikzlibrary{positioning,arrows.meta,fit,calc}
\usetikzlibrary[pgfplots.colormaps]
\DeclareUnicodeCharacter{2212}{−}
\usepgfplotslibrary{groupplots,dateplot}
\usetikzlibrary{patterns,shapes.arrows}
\pgfplotsset{compat=newest}
\usepackage{pgfplotstable}
\usepackage{siunitx}
\usepackage{booktabs} 
\usepackage{adjustbox}
\usepackage{multirow}
\usepackage{float}
\usepackage[nolist]{acronym}
\begin{acronym}
\acro{6G}{Sixth Generation}
\acro{AI}{Artificial Intelligence}
\acro{AP}{Average Precision}
\acro{AV}{Autonomous Vehicle}
\acro{AVS}{Autonomous Vehicles System}
\acro{CNN}{Convolutional Neural Network}
\acro{DNN}{Deep Neural Network}
\acro{G-PCC}{Geometry-Point Cloud Compression}
\acro{IoU}{Intersection over Union}
\acro{IoT}{Internet of Things}
\acro{LiDAR}{Light Detection and Ranging}
\acro{mAP}{mean Average Precision}
\acro{PV-RCNN}{PointVoxel-Region Convolutional Neural Network}
\acro{RPN}{Region Proposal Network}
\acro{RL}{Reinforcement Learning}
\acro{TD}{Teleoperated Driving}
\acro{V-PCC}{Video-Point Cloud Compression}
\acro{SELMA}{SEmantic Large Multimodal Acquisitions}
\acro{CV}{Computer Vision}
\acro{V2X}{Vehicle-to-Everything}
\acro{ES}{Extended Sensors}
\acro{e2e}{end-to-end}
\acro{PC}{Point Cloud}
\acro{M2M}{Machine to Machine}
\acro{KPI}{Key Performance Indicator}
\acro{SECOND}{Sparsely Embedded CONvolutional Detection}
\acro{3GPP}{3rd Generation Partnership Project}
\acro{E2E}{end-to-end}
\acro{UE}{User Equipment}
\acro{SUMO}{Simulation of Urban MObility}
\acro{GEMV2}{Geometry-based Efficient propagation Model for V2V communication}
\acro{SINR}{signal-to-interference-plus-noise ratio}
\acro{FR2}{Frequency Range 2}
\acro{OSM}{OpenStreetMap}
\end{acronym}

\usepackage[most]{tcolorbox}

\definecolor{title}{RGB}{255,180,154}

\newtcbtheorem{Summary}{\bfseries Summary}{enhanced,drop shadow={black!50!white},
  coltitle=black,
  top=0.3in,
  attach boxed title to top left=
  {xshift=1.5em,yshift=-\tcboxedtitleheight/2},
  boxed title style={size=small,colback=title}
}{summary}

\newtcolorbox[auto counter]{summary}[1][]{title={\bfseries Insight~\thetcbcounter},enhanced,drop shadow={black!50!white},
  coltitle=black,
  top=0.15in,
  attach boxed title to top left=
  {xshift=1.5em,yshift=-\tcboxedtitleheight/2},
  boxed title style={size=small,colback=title},#1}
\newtcolorbox[auto counter]{channel}[1][]{title={\bfseries Summary~\thetcbcounter},enhanced,drop shadow={black!50!white},
  coltitle=black,
  top=0.3in,
  attach boxed title to top left=
  {xshift=1.5em,yshift=-\tcboxedtitleheight/2},
  boxed title style={size=small,colback=title},#1}
  
\definecolor{red_cool}{rgb}{0.5, 0.0, 0.0}
\definecolor{orange}{rgb}{1, 0.5, 0.25}

\newcommand{\f}[1]{{\textbf{\textcolor{red}{f: #1}}}}

\definecolor{deep_blue}{rgb}{0.0,0.0,0.5}

\usepackage{flushend}

\begin{document}
\setlength{\extrarowheight}{2pt}
\title{Teleoperated Driving: a New Challenge for 3D Object Detection in Compressed Point Clouds}

\author{Filippo~Bragato~\orcidlink{0009-0006-1232-5402},~\IEEEmembership{Student Member, IEEE}, Michael~Neri~\orcidlink{0000-0002-6212-9139},~\IEEEmembership{Member~IEEE}, Paolo~Testolina~\orcidlink{0000-0002-5616-1722},  Marco~Giordani~\orcidlink{0000-0002-0575-1781},~\IEEEmembership{Senior Member, IEEE}, Federica~Battisti~\orcidlink{0000-0002-0846-5879},~\IEEEmembership{Senior Member, IEEE}
\thanks{F. Bragato, M. Giordani, and F. Battisti  are with the Department of Information Engineering, University of Padova. Padova, Italy. (e-mail: \{filippo.bragato,marco.giordani,federica.battisti\}@unipd.it). 
M. Neri is with the Faculty of Information Technology and Communication Sciences, Tampere University, Tampere, Finland (e-mail: {michael.neri@tuni.fi}). 
P. Testolina is with the Institute for Intelligent Networked Systems, Northeastern University, Boston, MA, USA (e-mail: {p.testolina@northeastern.edu}).
\textit{(Filippo Bragato and Michael Neri are co-first authors.)} }

}

\markboth{IEEE Transactions on Vehicular Technology,~Vol.~XX, No.~XX, October~2026}%
{Shell \MakeLowercase{\textit{et al.}}: A Sample Article Using IEEEtran.cls for IEEE Journals}


\maketitle
\begin{abstract}
In recent years, the development of interconnected devices has expanded in many fields, from infotainment to education and industrial applications. This trend has been accelerated by the increased number of low-cost sensors and accessibility to powerful hardware and software. One area that significantly benefits from these advancements is \ac{TD}. In this scenario, a remote controller (human or software) safely drives a vehicle using sensor data generated onboard the vehicle, and exchanged via Vehicle-to-Everything (V2X) communications. 
In this work, we tackle the problem of detecting the presence of cars, pedestrians, and cyclists from point cloud data generated by \ac{LiDAR} sensors to enable safe \ac{TD} operations. Specifically, we use the SELMA dataset, a multimodal, open-source, synthetic dataset for autonomous driving, that we expanded by including the ground-truth bounding boxes of 3D objects to support object detection. We analyze the performance of state-of-the-art compression algorithms and object detectors under several metrics, including compression efficiency, (de)compression and inference time, and detection accuracy. Finally, we measure the impact of compression and detection on a simulated V2X network in terms of data rate and latency with respect to the 3GPP requirements of \ac{TD} applications. Our results show that G-PCC outperforms its competitors in terms of both compression efficiency and accuracy, and consumes fewer channel resources for data transmissions. However, the substantial (de)compression time associated with its Octree representation prevents data processing within the typical latency requirements of \ac{TD} operations, regardless of the underlying compression configuration.
On the contrary, the two most conservative quantization modes of Draco, combined with SECOND or PointPillars for object detection, offer superior speed in
both compression and decompression with comparable accuracy. 
Finally, we show that the choice of the detector depends on the class of objects to detect, specifically pedestrians or~cyclists.
\end{abstract}

\begin{IEEEkeywords}
Point cloud compression; 3D object detection; G-PCC; Draco; Deep learning; SELMA.
\end{IEEEkeywords}
\acresetall

\section{Introduction}
\IEEEPARstart{I}n the last decades, the presence of sensors has expanded to almost every aspect of everyday life, from fitness and health wearables to vast \ac{IoT} networks. Sensor data is seen as a key enabler for a number of 6G applications, from cellular networks, where sensing is expected to play a central role, to video gaming, virtual reality, robotic navigation, and \ac{TD}~\cite{Neumeier_2019_ITSC, Chen_2021_MSP, Garrote_2022_Sensors}.

In the latter scenario, where safety, network, and control requirements are particularly strict, a remote controller (either a human or software agent) makes driving decisions for smart vehicles based on sensor data that captures the surrounding road environment. Notably, a suite of sensors is generally used to increase the reliability and accuracy of the digital representation of the scene. In this context, video cameras provide rich information, including color, in a format that is easy to interpret and process. However, they require adequate illumination and visibility conditions, compromising their reliability in dark, over or under exposed, or foggy scenarios~\cite{yue2018lidar}.

To mitigate these limitations, video cameras are often paired with \ac{LiDAR} and radar sensors. These sensors capture the characteristics of the environment by emitting an electromagnetic signal and measuring the backscattered radiation from nearby objects. This information is then represented through point clouds, i.e., discrete collections of data points represented by their spatial coordinates and, in some cases, additional attributes (e.g., the return signal intensity and color values)~\cite{li2020lidar}. In the automotive scenario, 3D point clouds can complement the 2D video camera data with depth information, and improve sensing performance even in adverse visibility and weather conditions~\cite{dai2024advanced}.

In addition, the transmission of large volumes of multimodal data generated by the sensor suite of the vehicles to an external processing unit is a promising solution for \ac{TD} applications~to:
\begin{enumerate}
    \item reduce the power consumption of (often electric) vehicles~\cite{traspadini2023realtime}, thus increasing their autonomy;
    \item aggregate data from multiple road users and sensors, obtaining a more complete view of the environment compared to the partial one available to each vehicle~\cite{11673199};
    \item employ larger processing models that cannot be deployed on the vehicle due to hardware constraints~\cite{Liu_2023_ScientificReports}.
\end{enumerate}
However, with raw data rates of several Gbps per vehicle, data transmissions via \ac{V2X} communication~\cite{zugno2020toward} can rapidly saturate the available network capacity, increase packet drop, and incur intolerable latency for safety-critical \ac{TD} operations. 
For example, a single raw \ac{LiDAR} point cloud is, on average, 1 MB~\cite{11421583}: at 30 fps, the data rate is as large as 240 Mbps.
 The fast-evolving network topology and channel propagation conditions in the road environment further complicate establishing a stable and reliable \ac{V2X} communication link~\cite{Feng_2021_TITS}. 


An important factor to optimize \ac{V2X} data transmission is the use of efficient compression algorithms~\cite{Garrote_2022_Sensors} to reduce the network load. However, compression can deteriorate data quality, which may reduce the accuracy of several \ac{TD} operations such as object detection and recognition. Furthermore, while compression is crucial to save network resources and accelerate transmissions, 
it introduces additional encoding and decoding delays that may
not be handled in real~time~\cite{nardo2022point}.

\begin{figure*}[t]
\centering
\includegraphics[width=0.99\linewidth]{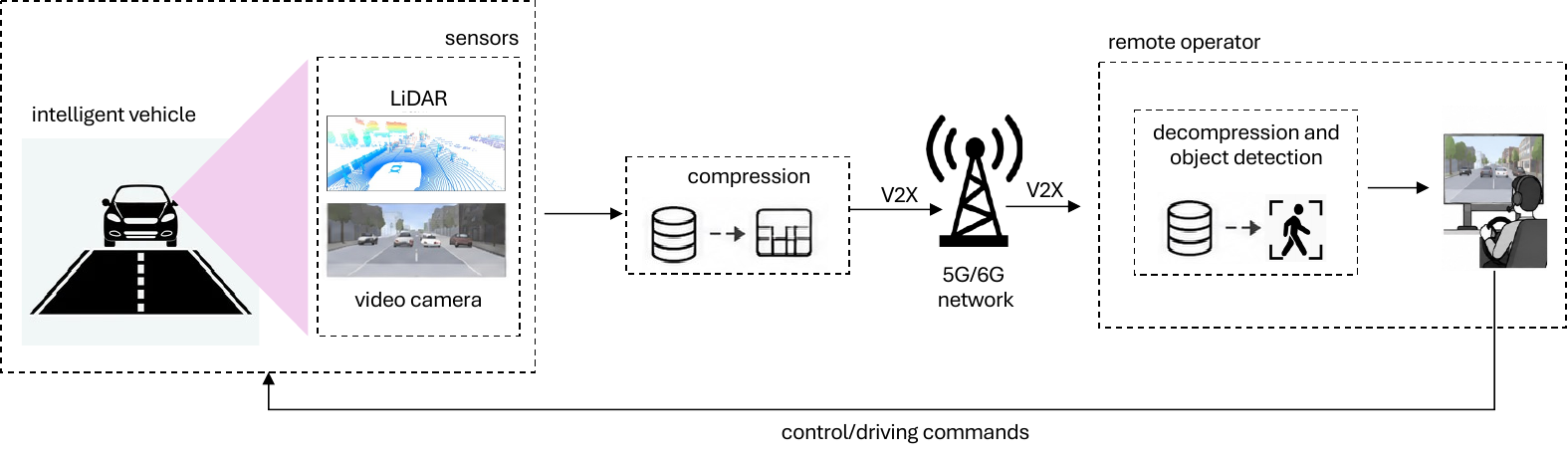}
\centering
\caption{Illustration of the reference TD scenario. Sensor data acquired by the vehicle are compressed and transmitted through the V2X network to a remote
processing unit, where they are reconstructed and used to support driving decisions. Control driving commands are subsequently transmitted back to the~vehicle.} 
\label{fig:scenario}
\end{figure*}

Motivated by these trade-offs, this paper investigates the question of \textit{how point cloud compression affects object detection and \ac{V2X} communication within a \ac{TD} scenario}.
Compared with prior studies, which typically investigate point cloud compression, object detection, or communication performance in isolation, we evaluate the complete data transmission and processing pipeline. 
To address this research, we consider the scenario in Fig.~\ref{fig:scenario}, where an intelligent vehicle generates, compresses, and transmits \ac{LiDAR} data to an external remote operator, deployed onboard a 5G/6G gNB, through the V2X network. The remote operator, in turn, performs object detection to detect the presence of cars, pedestrians, and cyclists on the received and uncompressed data. Driving commands are generated accordingly, and transmitted back to the~vehicle.
The feasibility of this scenario depends on the latency constraints of the \ac{TD} application, and the capacity of the underlying V2X network.
We provide the following~contributions:
\begin{itemize}
\item \ac{TD} algorithms like compressors and detectors are often based on complex deep learning models. To design and test these algorithms, realistic
automotive data is required. Therefore, we release a new version of the SELMA dataset~\cite{Testolina_2023_TITS}, a large-scale, multimodal, open-source synthetic automotive dataset, that includes ground truth 3D bounding boxes for object detection. 
Compared with existing automotive datasets, SELMA offers greater scenario diversity and denser annotations, particularly for vulnerable road users such as pedestrians and cyclists. 
Compression and detection algorithms are trained and validated on this dataset, ensuring fair comparison and reproducibility of the results.

\item We analyze the performance of two state-of-the-art codecs for \ac{LiDAR} data, namely \ac{G-PCC}~\cite{Graziosi_2020_APSIPA} and Draco~\cite{Draco_Github}, in terms of (de)compression speed and compression quality.
\item We compare the performance of three state-of-the-art object detectors, namely \ac{PV-RCNN}~\cite{Shi_2020_CVPR}, \ac{SECOND}~\cite{Yan_2018_Sensors}, and PointPillars~\cite{Lang_2019_CVPR}, on raw and compressed \ac{LiDAR} point clouds, in terms of \ac{AP} for cars, pedestrians, cyclists, and inference time.
\item We evaluate the combined impact of compression, decompression, inference, and transmission delays on the V2X network, and identify the codec–detector configurations that can satisfy capacity and delay constraints of the \ac{TD} application.
To do so, we run full-stack simulations using ns-3~\cite{riley2010ns} in terms of the data rate required to transmit raw and compressed \ac{LiDAR} data, and the resulting \ac{e2e} application delay.
\end{itemize}

We demonstrate that different \ac{LiDAR} compression and detection strategies offer different advantages depending on network constraints. 
Our results show that G-PCC outperforms its competitors in terms of both absolute compression efficiency and accuracy, and permits to use fewer channel resources for transmission. 
However, the Octree representation adopted by G-PCC incurs substantial time for both compression and inference, which prevents data processing and transmission within the typical latency requirements of \ac{TD} applications, set to 100 ms in this paper based on the 3GPP specifications~\cite{3gpp.22.186}, regardless of the underlying compression configuration.
In contrast, the two most conservative quantization modes of Draco, combined with SECOND or PointPillars, enable faster detection on the compressed data, and constitute a better solution for TD. 
Finally,  PointPillars achieves
higher~detection accuracy for pedestrians, while
SECOND is stronger on cyclists.

The rest of the paper is organized as follows. Sec.~\ref{sec:related} reviews the state-of-the-art on compression and detection algorithms applied to point clouds for TD applications. Sec.~\ref{sec:background} presents the compressors and detectors we selected for the experiments  in this paper. Sec.~\ref{sec:selma} describes the features of the extended SELMA dataset. Sec.~\ref{sec:results} presents our experimental results on compression and detection. Moreover, in Sec.~\ref{sec:network-results} we evaluate  the impact of both on the \ac{V2X} network, and provide guidelines for selecting the most suitable compression and detection algorithms based on the target constraint. In Sec.~\ref{sec:conclusions} we conclude the paper with suggestions for future~work.

\section{Related works}\label{sec:related}

Connected cars, when fully commercialized, will address the social and business trends of the next generation of transportation systems~\cite{boban2018connected}. In this regard, the \ac{3GPP}, since Release 15, has defined new use cases specific to future vehicular services. In this study, we consider the use case of \ac{TD}, and more specifically the scenario called \emph{``information sharing for automated driving between \acp{UE} supporting \ac{V2X} application''} in the 3GPP specifications~\cite{3gpp.22.186}. 
In this scenario,
automotive sensors, mainly video cameras and \acp{LiDAR}, provide a digital representation of the surrounding environment. Sensor data can be shared via \ac{V2X} communication with a remote controller, either a human operator or a software/server, sending driving commands to the vehicles actuators for proper control. 
In this scenario, the data rate is proportional to the resolution of the acquired data (typically less than $50$ Mbps), while delays must be very small (generally less than $100$ ms for high degrees of automation) to ensure prompt reactions to unpredictable events on the road, as outlined in~\cite{3gpp.22.186}.

Therefore, the size of the generated data, particularly point clouds from \ac{LiDAR} sensors, is a critical factor for \ac{TD} applications. Notably, point clouds can be compressed before data are broadcast to mitigate channel congestion and reduce data rates. At the time of writing, several methods have been proposed to compress point clouds while preserving quality. Given the 3D nature of LiDAR data, geometric compression algorithms, based on Point Cloud Data (PCD), LASComp/LASzip, and Octree formats, are the most common in the literature.
These methods require the use of deep learning techniques to either compress~\cite{Wiesmann_2021_LRA,Tu_2019_ICRA,Gao_2023_ICIP} or interpolate~\cite{Tu_2019_Access} the point clouds to reduce the number of points. Unfortunately, these codecs require point-level processing of data, which may not be implemented in real time onboard vehicles with limited computational capacity.

Furthermore, the quantification of the error introduced by compression is not trivial, and several metrics have been proposed in the literature~\cite{Lazzarotto_2021_MMSP}. In the context of \ac{TD}, the primary concern is the impact of compression on the performance of object detectors responsible for understanding the road scene. While the literature on 2D RGB images is quite mature~\cite{Gandor_2022_Sensors}, the effect of compression on 3D data is still an open challenge. For instance, the authors in~\cite{Wang_2023_TITS} investigated the impact of well-known video compression standards like Advanced Video Coding (AVC) and High Efficiency Video Coding (HEVC) on a state-of-the-art object detector like Faster R-CNN~\cite{Ren_2015_NeurIPS} in a vehicular scenario. A similar evaluation was done in~\cite{Garrote_2022_Sensors}, using \ac{G-PCC} with three levels of compression and PointNet++~\cite{Qi_2017_NeurIPS} as detector. However, the analysis involved converting the point clouds into images, and then applying 2D image-based techniques for compression, without directly operating on 3D~data.

\begin{figure*}[t]
\centering
\includegraphics[width=0.9\linewidth]{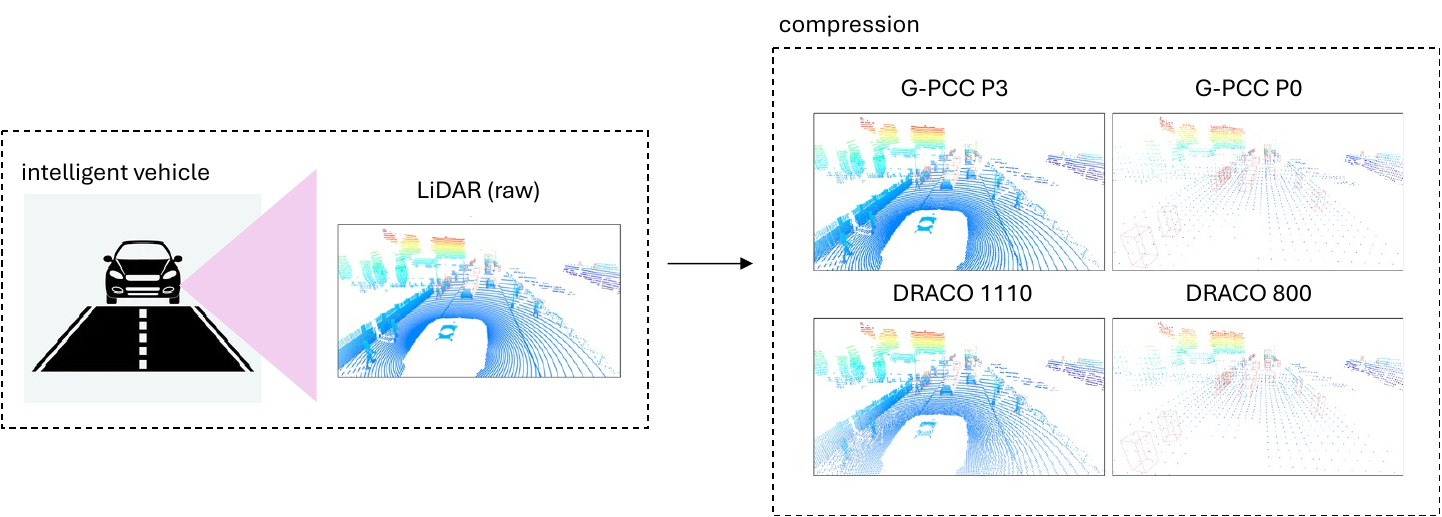}
\centering
\caption{Examples of a raw \ac{LiDAR} point cloud and its compressed versions using \ac{G-PCC} and Draco from the SELMA dataset. Ground truth bounding boxes are denoted in red.} \label{fig:pc_images}
\end{figure*}

The increasing attention to this topic has driven further research on the effects of compression on object detection in point clouds. For example, Martins \emph{et al.}~\cite{Martins2024} compared the effects of four codecs (a Cartesian-based JPEG Pleno PCC coder, a cylindrical coordinates based JPEG Pleno PCC coder, G-PCC, and L3C2) on the performance of four detectors (SECOND, PointPillars, PointRCNN and PV-RCNN). The analysis showed that point clouds can be compressed without significant loss of information. However, this study did not take into account the impact of the delay introduced by compression and detection on the \ac{V2X} network, which has been only partially explored in the literature. In~\cite{Varischio_2021_ICC} the effect of (de)compression time on the communication delay has been analyzed. However, a limited set of compression techniques has been considered, without assessing the impact of compression on the quality of object detectors. Similarly, in ~\cite{nardo2022point}, a comparison of 3D and 2D compression methods for point clouds, namely Octrees and G-PCC for 3D, and PNG, J-LS, LWZ and MJ2 for 2D, has been investigated. Compression quality was measured in terms of the Peak Signal-to-Noise Ratio (PSNR), which is, however, suboptimal for evaluating object detection performance unlike metrics such as the \ac{AP}.

It is important to highlight that most prior studies did not explore the impact of compression and detection on the \ac{V2X} communication network, which is essential for the proper design and dimensioning of \ac{TD} applications. A preliminary step in this direction was taken in~\cite{rossi2021role}, where the average data rate required to transmit sensor data (camera or LiDAR) using Aggregate View Object Detection (AVOD) has been estimated. However, the impact of point cloud compressors was not modeled, considering the \ac{V2X} channel and protocol stack ideal. 
To address these gaps, in this paper we compare different codecs and detection schemes, and evaluate network performance through complete, accurate, and realistic full-stack end-to-end V2X simulations using ns-3. 
We select the most mature, and widely adopted algorithms in the literature, which not only represent an established benchmark, but have also been used as starting points for the development of several more recent models, including the low-latency branches of G-PCC and emerging real-time learning codecs~\cite{you2025reno}.


\section{Background on Codecs and Object Detectors}
\label{sec:background}
This section provides insights regarding the codecs (Sec.~\ref{sub:compressors}) and object detectors (Sec.~\ref{sub:detectors}) under study. 

\subsection{Codecs}
\label{sub:compressors}
To analyze the effects of compression, two well-known point cloud compression methods are considered: \ac{G-PCC}~\cite{Graziosi_2020_APSIPA, Schwarz_2019_JETCAS_MPEG} and Draco~\cite{Draco_Github}. Examples of a raw point cloud and the output from \ac{G-PCC} and Draco are depicted in Fig.~\ref{fig:pc_images}. 

\subsubsection{G-PCC}
\ac{G-PCC}~\cite{Graziosi_2020_APSIPA, Schwarz_2019_JETCAS_MPEG} is developed by MPEG as part of the MPEG-I standard, and designed for lossless and lossy compression of 3D point clouds, preserving geometric structures with high accuracy. It employs techniques such as Octree partitioning, voxel-based coding, and adaptive quantization to efficiently encode spatial data. \ac{G-PCC} is particularly suited for LiDAR-based applications, as it can maintain fine geometric details, which is essential for object detection in \ac{TD} scenarios. It supports scalable bitrates, and poses trade-offs between compression efficiency and reconstruction fidelity, so it is adaptable to different bandwidth constraints.

We consider four different \ac{G-PCC} configurations, identified by label \textbf{pX}, where \textbf{X} is a number from $0$ to $3$ representing the  \verb|positionQuantizationScale| (PQS) parameter, which controls the number of quantization levels. We have:
\begin{itemize}
    \item {p0} (PQS $=0.0125$): high compression, low resolution;
    \item {p1} (PQS $=0.03125$);
    \item {p2} (PQS $=0.125$);
    \item {p3} (PQS $=0.375$): low compression, high resolution.
\end{itemize}

\subsubsection{Draco}
Draco~\cite{Draco_Github} is developed by Google, and it is a general-purpose 3D compression library optimized for both mesh and point cloud data. It uses predictive coding, quantization, and entropy coding to reduce file sizes while maintaining perceptual quality. Unlike G-PCC, Draco is primarily optimized for visualization and streaming applications, offering a balance between compression efficiency and fast decoding. Its lightweight design and lower computational complexity make it suitable for real-time applications where low-delay transmission of compressed point clouds is required.

Draco compression can be configured by two parameters, namely the quantization bits $q$ and the compression level $c$. 
Parameter $q$ represents the number of bits used for encoding the position of the points in the cloud: the higher $q$, the lower the error introduced by the compression, so the higher the size of the encoded point cloud. Parameter $c$ turns on and off different compression features, and generally a higher value corresponds to a lower size of the encoded cloud.
In this work, we consider twelve Draco configurations, identified by the expression $q\cdot100+c$, $\forall q \in \{8,9,10,11\}, \forall c \in \{0, 5, 10\}$, e.g., $905$ indicates $q=9$ and $c=5$.

\subsection{Object Detectors}
\label{sub:detectors}
State-of-the-art object detectors can be classified depending on the number of processing steps~\cite{Neri_2022_EUVIP}. First, we distinguish between single-stage and two-stage detectors:
\begin{itemize}
    \item {Single-stage detectors.} This class of detectors directly predicts object locations and class labels from the point cloud in a single pass. This approach promotes faster and more efficient object detection, and is suitable for real-time applications such as autonomous driving and robotics. Examples of single-stage detectors in the 3D domain include PointPillars~\cite{Lang_2019_CVPR}, which processes point clouds by projecting them into a pseudo-image for fast detection. While efficient, the accuracy of single-stage detectors may drop, especially in complex scenes or for smaller objects, without the refinement step.
    \item {Two-stage detectors.}  In the first stage, this class of detectors generates possible regions of interest where objects are likely to be located; then, in the second stage, these proposals are refined to improve the accuracy of both object localization and classification. A representative example is PV-RCNN~\cite{Shi_2020_CVPR}, which uses a voxel-based backbone for high-quality 3D proposals, and then refines them for more accurate detection. With respect to single-stage approaches, this two-stage paradigm usually requires much more computational resources, is slower, but performs better on complex scenes or objects that are difficult to detect.
\end{itemize}

In this work, we select the following and diverse object detectors: SECOND, PV-RCNN, and PointPillars. 
\ac{PV-RCNN}~\cite{Shi_2020_CVPR} is a two-stage detector, while PointPillars~\cite{Lang_2019_CVPR} and \ac{SECOND}~\cite{Yan_2018_Sensors} are single-stage detectors. \ac{SECOND} and \ac{PV-RCNN} are known for their strong detection performance, with \ac{PV-RCNN} leveraging point-based refinement for improved accuracy, while PointPillars~\cite{Lang_2019_CVPR} and \ac{SECOND} are lightweight, real-time alternatives for low-delay applications. These models process point clouds using varying degrees of voxelization and feature aggregation, making them ideal candidates to assess the impact of compression and the resulting sparsity and information loss. Moreover, all three detectors are widely used in vehicular scenarios~\cite{Martins2024}, and their performance under compressed point clouds directly translates into practical implications in terms of safety and efficiency for \ac{TD} applications. It is worth noting that several recent state-of-the-art 3D object detection frameworks build upon voxel- or pillar-based feature extractors derived from SECOND- or PointPillars-like backbones, extending them with transformer modules, temporal fusion, or improved feature aggregation strategies~\cite{zheng2025voxt, Chen_TITS_2025, cao2025kptr, zhou2023fastpillars}. As a result, evaluating compression effects on these representative detectors provides a practical lower-bound estimate on performance degradation and computational overhead that can be expected in a broader family of modern \ac{LiDAR} detection architectures used in autonomous and teleoperated driving systems.

\subsubsection{PV-RCNN}
\ac{PV-RCNN} is an object detection framework specifically tailored to 3D object detection in point clouds. It combines the advantages of point-based and voxel-based 3D detection methods, overcoming certain specific limitations of each individual approach~\cite{Shi_2020_CVPR}. \ac{PV-RCNN} initially converts the point cloud into a voxel representation, a process that involves transforming the continuous point cloud data into a structured, discrete grid. This voxelization step simplifies data processing, and is well-suited for convolutional neural networks, although it can sometimes lead to a loss of finer details. To compensate for this problem, \ac{PV-RCNN} also processes the raw point clouds in their original form to preserve intricate details, which is vital for accurate object detection in complex environments. 

However, the main drawback of \ac{PV-RCNN} is its complexity, thus the overall training and inference time. Although it is one of the most competitive object detectors in the KITTI benchmarks~\cite{Geiger_2013_IJRR}, its architecture is composed of tens of millions of parameters, complicating its use for real-time and resource-constrained applications.

\subsubsection{SECOND}
\ac{SECOND}~\cite{Yan_2018_Sensors} leverages sparse convolutions, which are efficient for processing the typically sparse 3D point cloud data generated by LiDAR sensors. It starts with the voxelization process, that is different from traditional methods as it focuses on keeping the representation sparse. The model then applies sparse 3D convolutions to the voxelized data. Unlike dense convolutions that process all voxels, sparse convolutions only focus on the non-empty voxels, significantly improving computational efficiency. In the later stages, SECOND employs a \ac{RPN} to generate 3D bounding box proposals for objects in the scene. As claimed in~\cite{Garrote_2022_Sensors}, this detector can hardly detect small objects in the 3D scene, such as pedestrians and cyclists. Similarly to voxelization-based detectors, this loss in performance can be due to the initial discretization step of the 3D environment.

\subsubsection{PointPillars}
PointPillars~\cite{Lang_2019_CVPR} initially organizes the unstructured point cloud data into a structured format known as ``pillar''. These pillars are essentially vertical columns in the point cloud, each encompassing a cluster of points. This pillar-based structure is significantly different from the traditional voxelization approach, offering a balance between preserving spatial details and computational efficiency. Then, the detector employs a neural network that operates on these pillars, extracting features from the raw point cloud data. The network is designed to handle the variable number of points in each pillar, ensuring that it captures the critical spatial information necessary for accurate object detection.

One of the core strengths of PointPillars is its ability to process point clouds in a way that is both computationally efficient and effective in retaining the vital spatial characteristics of the data. However, the pillar-based method may not capture the full intricacies of the object shapes and the spatial relationships in the environment, particularly in scenarios with complex geometries or highly cluttered scenes, e.g., a \ac{LiDAR} scan in a very crowded area~\cite{Stanisz_2020_SPA}.

\begin{figure*}[t]
\centering
\includegraphics[width=0.95\linewidth]{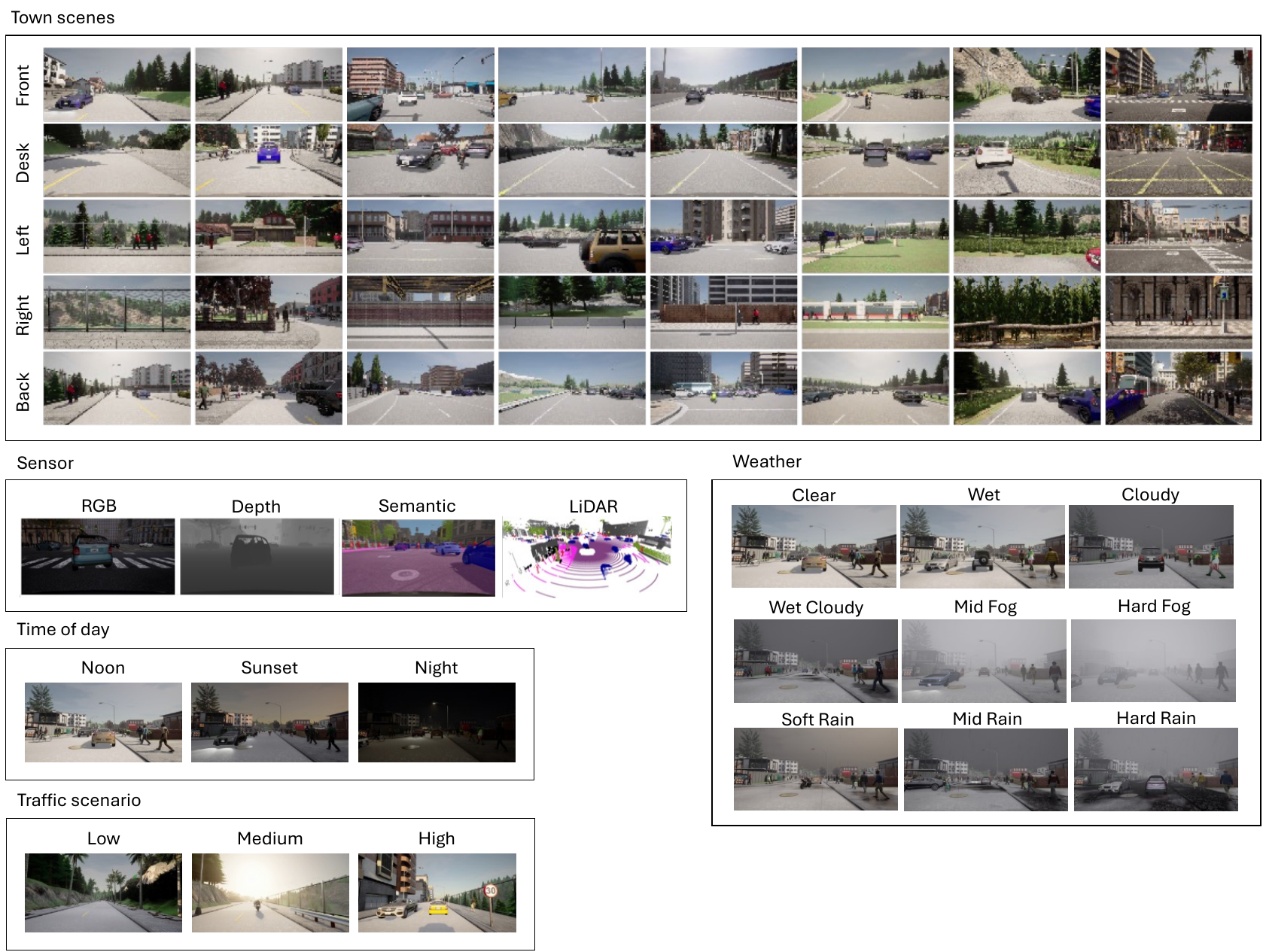}
\centering
\caption{Randomly sampled images from the SELMA dataset in different multimodal conditions.} \label{fig:selma}
\end{figure*}

\begin{table}[t!]
    \centering
    \caption{Comparison of SELMA with other state-of-the-art datasets.}
    \label{tab:dataset_comparison}
    \begin{tabular}{c|ccc}
        \toprule
                              & \textbf{KITTI}~\cite{Geiger_2013_IJRR} & \textbf{Waymo}~\cite{Sun_2020_Waymo}           & \textbf{SELMA}~\cite{Testolina_2023_TITS} \\
        \midrule
        Frames             & $15$K            & $230$K                     & $2.5$M           \\
        3D boxes           & $430$K           & $12$M                      & $150$M           \\
        \midrule 
        LiDAR(s)             & $1$              & $5$                        & $3$              \\
        Range                 & \SI{120}{m}    & \{75,100\} m  & \SI{200}{m}   \\
        Points             & $120$K           & $177$K                         & $91$K            \\
        \midrule
        Vehicles           & $180$K           & $6.1$M                     & $68$M            \\
        Pedestrians        & $110$K           & $2.8$M                     & $61$M            \\
        Cyclists           & $20$K            & $67$K                      & $18$M            \\
        \midrule 
        Vehicles/frame    & $12$             & $26.5$                     & $32.9$           \\
        Pedestrians/frame & $7.3$            & $12.2$                     & $28.2$           \\
        Cyclists/frame    & $1.3$            & $0.3$                      & $8.5$            \\
        \bottomrule 
    \end{tabular}
\end{table}

\section{Extended SELMA Dataset}
\label{sec:selma}
In this section we describe the dataset that we will use in the experiments described in Secs.~\ref{sec:results} and \ref{sec:network-results}, which is an extension of the SELMA dataset presented in~\cite{Testolina_2023_TITS}.\footnote{SELMA is available at \url{https://scanlab.dei.unipd.it/selma-dataset/}.} 

\ac{SELMA} is a multimodal, open-source, synthetic dataset for autonomous driving. It was generated using the CARLA simulator \cite{CARLA_2017}, and consists of data acquired in $30\,909$ independent locations from $7$ RGB semantic cameras, $7$ depth cameras, and $3$ LiDARs, for a total of $2.5$ million frames. Notably, SELMA is one of the few open-source datasets to provide labeled data for multiple and diverse traffic scenarios, generated from different viewpoints, weather, lighting, and daytime conditions, for a total of $216$ unique settings, as illustrated in Fig.~\ref{fig:selma}. Moreover, it supports the full class set of common benchmarks like Cityscapes~\cite{cordts2016cityscapes}. The multimodal nature of SELMA promotes data complementarity and diversity, and baseline experiments proved that it achieves better performance, e.g., in terms of semantic segmentation accuracy, both in the real domain and against competing synthetic datasets~\cite{Testolina_2023_TITS}. 

In its original version, \ac{SELMA} was designed for the evaluation of semantic segmentation tasks. In particular, the data from each sensor was annotated with semantic labels for each pixel (for cameras) and point (for LiDARs). In this work, we extended the dataset to also support the evaluation of object detection tasks by providing bounding boxes for each object in the scenes. 
Specifically, bounding boxes include the position, orientation, and dimensions of each object, and have been generated automatically using a modified version of the CARLA simulation engine. This approach eliminates human labeling errors, inconsistencies, or boundary noise which may be possibly introduced via manual annotation~\cite{CARLA_2017}.


In Table~\ref{tab:dataset_comparison} we compare the \ac{SELMA} dataset against other state-of-the-art competitors, such as KITTI~\cite{Geiger_2013_IJRR} and Waymo~\cite{Sun_2020_Waymo}. \ac{SELMA} appears as the largest dataset in terms of the number of frames and 3D bounding boxes, and comes with a higher number of vehicles, pedestrians, and cyclists per frame compared to the other datasets.
Overall, the dataset includes $150$ million 3D bounding boxes for $68$ million vehicles, $61$ million pedestrians, and $18$ million cyclists, with an average of $28.2$ pedestrians and $8.5$ cyclists per frame, compared to $12.2$ pedestrians and $0.3$ cyclists per frame in the Waymo dataset~\cite{Sun_2020_Waymo}. This high-density annotation strategy is intended to improve the detection of cyclists and pedestrians, who are the most vulnerable road users.

\section{Experimental Results for Compression and Detection Performance}\label{sec:results}
In this section we present our simulation parameters (Sec.~\ref{sec:training_details}) and metrics (Sec.~\ref{sub:metrics}), and evaluate the impact of compression (Sec.~\ref{sub:compression-results}) and object detection (Sec.~\ref{sub:detection-performance}) on a reference \ac{TD} application sending and processing LiDAR~data.

\subsection{Simulation Parameters} \label{sec:training_details}

Training and testing of the codecs and detectors have been carried out on a workstation with $64$ GB of RAM and a NVIDIA GeForce RTX 3090 Ti graphics card. We used the \ac{SELMA} dataset presented in Sec.~\ref{sec:selma}, considering LiDAR point clouds of $91$K points on average. For object detection, all the models have been trained using an open-source toolbox based on PyTorch\footnote{\url{https://github.com/open-mmlab/mmdetection3d}} with $80$ epochs, a cosine annealing learning rate of $0.001$ with period $T_{\mathrm{max}} = 48$, and a batch size of $4$ for SECOND and PointPillars and $2$ for PV-RCNN. We follow the same training-validation-testing split proposed in~\cite{Testolina_2023_TITS}, that is a random $80$-$10$-$10$ split. 
For G-PCC, we use the MPEG TMC13 reference implementation~\cite{liu2019comprehensive}.
Notice that all detectors are trained on uncompressed point clouds and evaluated on compressed ones. 
This experimental setting captures two complementary effects, namely the information loss introduced by compression and the distribution shift between uncompressed and compressed point clouds. Nevertheless, retraining or fine-tuning the detectors on compressed data could likely mitigate part of these degradation effects.

\subsection{Evaluation Metrics}
\label{sub:metrics}
We evaluate the compression efficiency, measured as the capability of the codec to reduce the file size of the point cloud, and the compression (encoding) and decompression (decoding) time, measured from the moment at which the raw point cloud is produced until the compressed point cloud is generated or vice versa,~respectively.

For object detection, we measure the quality of each model in terms of the \ac{AP} for each class of objects~\cite{everingham2010pascal}. It is related to the precision-recall curve, where precision (recall) is defined as the percentage of correct predictions (ground-truth objects) over all the predictions (ground-truth objects), ranked above a given threshold.  The \ac{AP} is defined as the average of the interpolated precision values over a set $\mathcal{R}$ of equally-spaced recall thresholds,\footnote{In this study, we consider forty thresholds, i.e., $\{0.025,0.05,\dots,1\}$, as proposed in the KITTI dataset~\cite{Geiger_2013_IJRR}.} that is 
\begin{equation}
    \mathrm{AP} = {1}/{|\mathcal{R}|} \sum_{r\in\mathcal{R}} P_i(r),
\end{equation}
 where $P_i(r)$ is the interpolated precision at recall level $r$, i.e.,
\begin{equation}\label{interpo}
   P_i(r) = \max_{\tilde{r} : \tilde{r} \ge r} P(\tilde{r}),
\end{equation}
where $P(\tilde{r})$ denotes the precision at recall $\tilde{r}$. This interpolation ensures that the precision at each recall level $r$ is the highest precision obtained for any recall greater than or equal to $r$. 

\subsection{Compression Results}
\label{sub:compression-results}
We evaluate the performance of the codecs described in Sec.~\ref{sub:compressors} in terms of the size of the compressed point clouds and the (de)compression time. Specifically, we consider four configurations for G-PCC and twelve for Draco, to trade-off efficiency and quality.

\begin{figure}[!tbp]
  \centering
\pgfplotsset{
tick label style={font=\footnotesize},
label style={font=\footnotesize},
legend  style={font=\footnotesize}
}
\begin{tikzpicture}

\definecolor{crimson2233131}{RGB}{48,103,141}
\definecolor{darkgray176}{RGB}{176,176,176}
\definecolor{darkslategray76}{RGB}{76,76,76}
\definecolor{mediumblue3131223}{RGB}{53,183,120}

\begin{groupplot}[group style={group size=2 by 1, horizontal sep=0.1cm}]
\nextgroupplot[
tick align=outside,
tick pos=left,
x grid style={darkgray176},
xlabel={Draco},
xmin=-0.5, xmax=11.5,
xtick style={color=black},
xtick={0,1,2,3,4,5,6,7,8,9,10,11},
xticklabels={800,805,810,900,905,910,1000,1005,1010,1100,1105,1110},
xticklabel style={rotate=90},
y grid style={darkgray176},
ymajorgrids,
yminorgrids,
ylabel={Compressed file size (KB)},
ymin=-4.0934, ymax=90.9114,
ytick style={color=black},
width=0.75\linewidth,
height=0.65\linewidth
]
\path [draw=darkslategray76, fill=crimson2233131, semithick]
(axis cs:-0.4,11.179)
--(axis cs:0.4,11.179)
--(axis cs:0.4,18.033)
--(axis cs:-0.4,18.033)
--(axis cs:-0.4,11.179)
--cycle;
\path [draw=darkslategray76, fill=crimson2233131, semithick]
(axis cs:0.6,4.761)
--(axis cs:1.4,4.761)
--(axis cs:1.4,8.663)
--(axis cs:0.6,8.663)
--(axis cs:0.6,4.761)
--cycle;
\path [draw=darkslategray76, fill=crimson2233131, semithick]
(axis cs:1.6,4.826)
--(axis cs:2.4,4.826)
--(axis cs:2.4,8.7565)
--(axis cs:1.6,8.7565)
--(axis cs:1.6,4.826)
--cycle;
\path [draw=darkslategray76, fill=crimson2233131, semithick]
(axis cs:2.6,16.721)
--(axis cs:3.4,16.721)
--(axis cs:3.4,26.554)
--(axis cs:2.6,26.554)
--(axis cs:2.6,16.721)
--cycle;
\path [draw=darkslategray76, fill=crimson2233131, semithick]
(axis cs:3.6,8.9135)
--(axis cs:4.4,8.9135)
--(axis cs:4.4,16.391)
--(axis cs:3.6,16.391)
--(axis cs:3.6,8.9135)
--cycle;
\path [draw=darkslategray76, fill=crimson2233131, semithick]
(axis cs:4.6,8.608)
--(axis cs:5.4,8.608)
--(axis cs:5.4,15.549)
--(axis cs:4.6,15.549)
--(axis cs:4.6,8.608)
--cycle;
\path [draw=darkslategray76, fill=crimson2233131, semithick]
(axis cs:5.6,24.929)
--(axis cs:6.4,24.929)
--(axis cs:6.4,39.1705)
--(axis cs:5.6,39.1705)
--(axis cs:5.6,24.929)
--cycle;
\path [draw=darkslategray76, fill=crimson2233131, semithick]
(axis cs:6.6,15.8095)
--(axis cs:7.4,15.8095)
--(axis cs:7.4,28.98)
--(axis cs:6.6,28.98)
--(axis cs:6.6,15.8095)
--cycle;
\path [draw=darkslategray76, fill=crimson2233131, semithick]
(axis cs:7.6,14.516)
--(axis cs:8.4,14.516)
--(axis cs:8.4,25.7435)
--(axis cs:7.6,25.7435)
--(axis cs:7.6,14.516)
--cycle;
\path [draw=darkslategray76, fill=crimson2233131, semithick]
(axis cs:8.6,36.559)
--(axis cs:9.4,36.559)
--(axis cs:9.4,56.5955)
--(axis cs:8.6,56.5955)
--(axis cs:8.6,36.559)
--cycle;
\path [draw=darkslategray76, fill=crimson2233131, semithick]
(axis cs:9.6,26.44)
--(axis cs:10.4,26.44)
--(axis cs:10.4,47.3715)
--(axis cs:9.6,47.3715)
--(axis cs:9.6,26.44)
--cycle;
\path [draw=darkslategray76, fill=crimson2233131, semithick]
(axis cs:10.6,23.0655)
--(axis cs:11.4,23.0655)
--(axis cs:11.4,40.228)
--(axis cs:10.6,40.228)
--(axis cs:10.6,23.0655)
--cycle;
\addplot [semithick, darkslategray76]
table {%
0 11.179
0 3.249
};
\addplot [semithick, darkslategray76]
table {%
0 18.033
0 28.314
};
\addplot [semithick, darkslategray76]
table {%
-0.2 3.249
0.2 3.249
};
\addplot [semithick, darkslategray76]
table {%
-0.2 28.314
0.2 28.314
};
\addplot [semithick, darkslategray76]
table {%
1 4.761
1 0.85
};
\addplot [semithick, darkslategray76]
table {%
1 8.663
1 14.516
};
\addplot [semithick, darkslategray76]
table {%
0.8 0.85
1.2 0.85
};
\addplot [semithick, darkslategray76]
table {%
0.8 14.516
1.2 14.516
};
\addplot [semithick, darkslategray76]
table {%
2 4.826
2 0.852
};
\addplot [semithick, darkslategray76]
table {%
2 8.7565
2 14.641
};
\addplot [semithick, darkslategray76]
table {%
1.8 0.852
2.2 0.852
};
\addplot [semithick, darkslategray76]
table {%
1.8 14.641
2.2 14.641
};
\addplot [semithick, darkslategray76]
table {%
3 16.721
3 4.736
};
\addplot [semithick, darkslategray76]
table {%
3 26.554
3 41.291
};
\addplot [semithick, darkslategray76]
table {%
2.8 4.736
3.2 4.736
};
\addplot [semithick, darkslategray76]
table {%
2.8 41.291
3.2 41.291
};
\addplot [semithick, darkslategray76]
table {%
4 8.9135
4 1.613
};
\addplot [semithick, darkslategray76]
table {%
4 16.391
4 27.536
};
\addplot [semithick, darkslategray76]
table {%
3.8 1.613
4.2 1.613
};
\addplot [semithick, darkslategray76]
table {%
3.8 27.536
4.2 27.536
};
\addplot [semithick, darkslategray76]
table {%
5 8.608
5 1.566
};
\addplot [semithick, darkslategray76]
table {%
5 15.549
5 25.935
};
\addplot [semithick, darkslategray76]
table {%
4.8 1.566
5.2 1.566
};
\addplot [semithick, darkslategray76]
table {%
4.8 25.935
5.2 25.935
};
\addplot [semithick, darkslategray76]
table {%
6 24.929
6 7.18
};
\addplot [semithick, darkslategray76]
table {%
6 39.1705
6 60.526
};
\addplot [semithick, darkslategray76]
table {%
5.8 7.18
6.2 7.18
};
\addplot [semithick, darkslategray76]
table {%
5.8 60.526
6.2 60.526
};
\addplot [semithick, darkslategray76]
table {%
7 15.8095
7 3.094
};
\addplot [semithick, darkslategray76]
table {%
7 28.98
7 47.833
};
\addplot [semithick, darkslategray76]
table {%
6.8 3.094
7.2 3.094
};
\addplot [semithick, darkslategray76]
table {%
6.8 47.833
7.2 47.833
};
\addplot [semithick, darkslategray76]
table {%
8 14.516
8 2.776
};
\addplot [semithick, darkslategray76]
table {%
8 25.7435
8 42.461
};
\addplot [semithick, darkslategray76]
table {%
7.8 2.776
8.2 2.776
};
\addplot [semithick, darkslategray76]
table {%
7.8 42.461
8.2 42.461
};
\addplot [semithick, darkslategray76]
table {%
9 36.559
9 10.919
};
\addplot [semithick, darkslategray76]
table {%
9 56.5955
9 86.593
};
\addplot [semithick, darkslategray76]
table {%
8.8 10.919
9.2 10.919
};
\addplot [semithick, darkslategray76]
table {%
8.8 86.593
9.2 86.593
};
\addplot [semithick, darkslategray76]
table {%
10 26.44
10 5.842
};
\addplot [semithick, darkslategray76]
table {%
10 47.3715
10 74.333
};
\addplot [semithick, darkslategray76]
table {%
9.8 5.842
10.2 5.842
};
\addplot [semithick, darkslategray76]
table {%
9.8 74.333
10.2 74.333
};
\addplot [semithick, darkslategray76]
table {%
11 23.0655
11 5.078
};
\addplot [semithick, darkslategray76]
table {%
11 40.228
11 65.622
};
\addplot [semithick, darkslategray76]
table {%
10.8 5.078
11.2 5.078
};
\addplot [semithick, darkslategray76]
table {%
10.8 65.622
11.2 65.622
};
\addplot [semithick, darkslategray76]
table {%
-0.4 13.711
0.4 13.711
};
\addplot [semithick, darkslategray76]
table {%
0.6 6.15
1.4 6.15
};
\addplot [semithick, darkslategray76]
table {%
1.6 6.226
2.4 6.226
};
\addplot [semithick, darkslategray76]
table {%
2.6 20.231
3.4 20.231
};
\addplot [semithick, darkslategray76]
table {%
3.6 11.323
4.4 11.323
};
\addplot [semithick, darkslategray76]
table {%
4.6 10.972
5.4 10.972
};
\addplot [semithick, darkslategray76]
table {%
5.6 29.879
6.4 29.879
};
\addplot [semithick, darkslategray76]
table {%
6.6 19.482
7.4 19.482
};
\addplot [semithick, darkslategray76]
table {%
7.6 18.069
8.4 18.069
};
\addplot [semithick, darkslategray76]
table {%
8.6 43.44
9.4 43.44
};
\addplot [semithick, darkslategray76]
table {%
9.6 31.452
10.4 31.452
};
\addplot [semithick, darkslategray76]
table {%
10.6 27.901
11.4 27.901
};

\nextgroupplot[
scaled y ticks=manual:{}{\pgfmathparse{#1}},
tick align=outside,
tick pos=left,
x grid style={darkgray176},
xlabel={G-PCC},
xmin=-0.5, xmax=3.5,
xtick style={color=black},
xtick={0,1,2,3},
xticklabels={p0,p1,p2,p3},
y grid style={darkgray176},
ymin=-4.0934, ymax=90.9114,
ytick style={color=black},
yticklabels={},
ymajorgrids,
yminorgrids,
width=0.4\linewidth,
height=0.65\linewidth
]
\path [draw=darkslategray76, fill=mediumblue3131223, semithick]
(axis cs:-0.4,1.305)
--(axis cs:0.4,1.305)
--(axis cs:0.4,2.438)
--(axis cs:-0.4,2.438)
--(axis cs:-0.4,1.305)
--cycle;
\path [draw=darkslategray76, fill=mediumblue3131223, semithick]
(axis cs:0.6,3.394)
--(axis cs:1.4,3.394)
--(axis cs:1.4,6.447)
--(axis cs:0.6,6.447)
--(axis cs:0.6,3.394)
--cycle;
\path [draw=darkslategray76, fill=mediumblue3131223, semithick]
(axis cs:1.6,11.237)
--(axis cs:2.4,11.237)
--(axis cs:2.4,20.883)
--(axis cs:1.6,20.883)
--(axis cs:1.6,11.237)
--cycle;
\path [draw=darkslategray76, fill=mediumblue3131223, semithick]
(axis cs:2.6,24.748)
--(axis cs:3.4,24.748)
--(axis cs:3.4,45.377)
--(axis cs:2.6,45.377)
--(axis cs:2.6,24.748)
--cycle;
\addplot [semithick, darkslategray76]
table {%
0 1.305
0 0.225
};
\addplot [semithick, darkslategray76]
table {%
0 2.438
0 4.133
};
\addplot [semithick, darkslategray76]
table {%
-0.2 0.225
0.2 0.225
};
\addplot [semithick, darkslategray76]
table {%
-0.2 4.133
0.2 4.133
};
\addplot [semithick, darkslategray76]
table {%
1 3.394
1 0.445
};
\addplot [semithick, darkslategray76]
table {%
1 6.447
1 11.025
};
\addplot [semithick, darkslategray76]
table {%
0.8 0.445
1.2 0.445
};
\addplot [semithick, darkslategray76]
table {%
0.8 11.025
1.2 11.025
};
\addplot [semithick, darkslategray76]
table {%
2 11.237
2 1.839
};
\addplot [semithick, darkslategray76]
table {%
2 20.883
2 35.183
};
\addplot [semithick, darkslategray76]
table {%
1.8 1.839
2.2 1.839
};
\addplot [semithick, darkslategray76]
table {%
1.8 35.183
2.2 35.183
};
\addplot [semithick, darkslategray76]
table {%
3 24.748
3 5.788
};
\addplot [semithick, darkslategray76]
table {%
3 45.377
3 68.462
};
\addplot [semithick, darkslategray76]
table {%
2.8 5.788
3.2 5.788
};
\addplot [semithick, darkslategray76]
table {%
2.8 68.462
3.2 68.462
};
\addplot [semithick, darkslategray76]
table {%
-0.4 1.726
0.4 1.726
};
\addplot [semithick, darkslategray76]
table {%
0.6 4.459
1.4 4.459
};
\addplot [semithick, darkslategray76]
table {%
1.6 14.097
2.4 14.097
};
\addplot [semithick, darkslategray76]
table {%
2.6 29.162
3.4 29.162
};
\end{groupplot}

\end{tikzpicture}
\vspace{-30pt}
  \caption{Compressed file size of the point cloud vs. the compression configuration of G-PCC and Draco.}
  \label{fig:comparison_compressors}
\end{figure}

\subsubsection{Compressed file size}

In Fig.~\ref{fig:comparison_compressors} we illustrate the impact of the compression configuration on the encoded file size. For comparison, the raw size of a single uncompressed point cloud frame is about $1066$ KB. Regarding G-PCC, the performance depends on the PQS parameter, identified by labels \{p0, p1, p2, p3\}, which is related to the number of bits in the point cloud. Therefore, the resulting file size after compression decreases as PQS decreases. Notably, the variance of the compressed file size increases for lower compression configurations (i.e., p2 and p3), whereas it is more stable as compression increases. In contrast, for Draco, the compressed file size increases with $q$ and decreases with $c$. As such, Draco is more sensitive to the number of quantization bits in the resulting point cloud than the compression configurations, particularly when $q$ is small. 
In terms of absolute performance, G-PCC outperforms Draco: configuration p0 produces files of $1.9$ KB on average, against $6.8$ KB for the smallest Draco modes ($805$ and $810$), while at the opposite end p3 results in an average file size of $33.4$ KB vs. $37.6$ KB for the $11$xx~group.

\begin{figure}[t!]
  \centering
\pgfplotsset{
tick label style={font=\footnotesize},
label style={font=\footnotesize},
legend  style={font=\footnotesize}
}
\begin{tikzpicture}

\definecolor{darkgray153}{RGB}{153,153,153}
\definecolor{darkgray176}{RGB}{176,176,176}
\definecolor{darkslategray76}{RGB}{76,76,76}
\definecolor{l1}{RGB}{64,67,135}
\definecolor{l2}{RGB}{41,120,142}
\definecolor{l3}{RGB}{34,167,132}
\definecolor{l4}{RGB}{121,209,81}

\begin{axis}[
tick align=outside,
tick pos=left,
x grid style={darkgray176},
xmin=-0.5, xmax=3.5,
xlabel={G-PCC compression configuration},
xtick style={color=black},
xtick={0,1,2,3},
xticklabels={p0,p1,p2,p3},
y grid style={darkgray176},
ymajorgrids,
yminorgrids,
ylabel={Compression + decompression time (ms)},
ymin=24.8, ymax=1063.2,
ytick style={color=black},
width=\linewidth,
height=0.65\linewidth
]
\path [draw=darkgray153, fill=l1, semithick]
(axis cs:-0.4,96)
--(axis cs:0.4,96)
--(axis cs:0.4,116)
--(axis cs:-0.4,116)
--(axis cs:-0.4,96)
--cycle;
\path [draw=darkgray153, fill=l2, semithick]
(axis cs:0.6,134)
--(axis cs:1.4,134)
--(axis cs:1.4,187)
--(axis cs:0.6,187)
--(axis cs:0.6,134)
--cycle;
\path [draw=darkgray153, fill=l3, semithick]
(axis cs:1.6,275)
--(axis cs:2.4,275)
--(axis cs:2.4,450)
--(axis cs:1.6,450)
--(axis cs:1.6,275)
--cycle;
\path [draw=darkgray153, fill=l4, semithick]
(axis cs:2.6,501)
--(axis cs:3.4,501)
--(axis cs:3.4,819)
--(axis cs:2.6,819)
--(axis cs:2.6,501)
--cycle;
\addplot [semithick, darkgray153]
table {%
0 96
0 72
};
\addplot [semithick, darkgray153]
table {%
0 116
0 146
};
\addplot [semithick, darkgray153]
table {%
-0.2 72
0.2 72
};
\addplot [semithick, darkgray153]
table {%
-0.2 146
0.2 146
};
\addplot [semithick, darkgray153]
table {%
1 134
1 82
};
\addplot [semithick, darkgray153]
table {%
1 187
1 266
};
\addplot [semithick, darkgray153]
table {%
0.8 82
1.2 82
};
\addplot [semithick, darkgray153]
table {%
0.8 266
1.2 266
};
\addplot [semithick, darkgray153]
table {%
2 275
2 133
};
\addplot [semithick, darkgray153]
table {%
2 450
2 625
};
\addplot [semithick, darkgray153]
table {%
1.8 133
2.2 133
};
\addplot [semithick, darkgray153]
table {%
1.8 625
2.2 625
};
\addplot [semithick, darkgray153]
table {%
3 501
3 211
};
\addplot [semithick, darkgray153]
table {%
3 819
3 1016
};
\addplot [semithick, darkgray153]
table {%
2.8 211
3.2 211
};
\addplot [semithick, darkgray153]
table {%
2.8 1016
3.2 1016
};
\addplot [semithick, darkgray153]
table {%
-0.4 103
0.4 103
};
\addplot [semithick, darkgray153]
table {%
0.6 151
1.4 151
};
\addplot [semithick, darkgray153]
table {%
1.6 314
2.4 314
};
\addplot [semithick, darkgray153]
table {%
2.6 550
3.4 550
};
\end{axis}

\end{tikzpicture}
\vspace{-40pt}
    \caption{Total compression and decompression time vs. the compression configuration (in terms of the PQS parameter) for G-PCC.}
    \label{fig:gpcc_compression_params}
\end{figure}

\begin{figure}[t!]
  \centering
  \input{images/fig_codec_time_size}
  \caption{Compression and decompression time, and the sum of both vs. the compressed file size for G-PCC (top) and Draco (bottom). Note the different vertical scales. Decoding is faster than encoding, which is particularly important for \ac{TD} applications to accelerate the execution of subsequent processing tasks at the receiver.}
  \label{fig:codec_time_size}
\end{figure}

\begin{figure}[t!]
  \centering
\pgfplotsset{
tick label style={font=\footnotesize},
label style={font=\footnotesize},
legend  style={font=\footnotesize}
}
\begin{tikzpicture}

\definecolor{aqua0255245}{RGB}{0,255,245}
\definecolor{chartreuse1312550}{RGB}{131,255,0}
\definecolor{darkgray176}{RGB}{176,176,176}
\definecolor{darkslategray76}{RGB}{76,76,76}
\definecolor{darkviolet1130255}{RGB}{113,0,255}
\definecolor{gray144}{RGB}{144,144,144}
\definecolor{lightgray204}{RGB}{204,204,204}
\definecolor{c0}{RGB}{240,240,240}
\definecolor{c5}{RGB}{247,247,247}
\definecolor{c0}{RGB}{58,82,139}
\definecolor{c5}{RGB}{32,144,140}
\definecolor{c10}{RGB}{94,201,97}

\begin{axis}[
legend cell align={left},
legend style={
  fill opacity=0.8,
  draw opacity=1,
  text opacity=1,
  at={(0.03,0.97)},
  anchor=north west,
  draw=lightgray204
},
tick align=outside,
tick pos=left,
x grid style={darkgray176},
xlabel={Draco quantization bits ($q$)},
xmin=-0.5, xmax=3.5,
xtick style={color=black},
xtick={0,1,2,3},
xticklabels={8,9,10,11},
y grid style={darkgray176},
ymajorgrids,
yminorgrids,
ylabel={Compression + decompression time (ms)},
ymin=4.95, ymax=28.05,
ytick style={color=black},
width=\linewidth,
height=0.65\linewidth
]
\path [draw=gray144, fill=c0, semithick]
(axis cs:-0.397333333333333,11)
--(axis cs:-0.136,11)
--(axis cs:-0.136,13)
--(axis cs:-0.397333333333333,13)
--(axis cs:-0.397333333333333,11)
--cycle;
\path [draw=gray144, fill=c5, semithick]
(axis cs:-0.130666666666667,8)
--(axis cs:0.130666666666667,8)
--(axis cs:0.130666666666667,11)
--(axis cs:-0.130666666666667,11)
--(axis cs:-0.130666666666667,8)
--cycle;
\path [draw=gray144, fill=c10, semithick]
(axis cs:0.136,11)
--(axis cs:0.397333333333333,11)
--(axis cs:0.397333333333333,14)
--(axis cs:0.136,14)
--(axis cs:0.136,11)
--cycle;
\path [draw=gray144, fill=c0, semithick]
(axis cs:0.602666666666667,10)
--(axis cs:0.864,10)
--(axis cs:0.864,13)
--(axis cs:0.602666666666667,13)
--(axis cs:0.602666666666667,10)
--cycle;
\path [draw=gray144, fill=c5, semithick]
(axis cs:0.869333333333333,9)
--(axis cs:1.13066666666667,9)
--(axis cs:1.13066666666667,12)
--(axis cs:0.869333333333333,12)
--(axis cs:0.869333333333333,9)
--cycle;
\path [draw=gray144, fill=c10, semithick]
(axis cs:1.136,13)
--(axis cs:1.39733333333333,13)
--(axis cs:1.39733333333333,16)
--(axis cs:1.136,16)
--(axis cs:1.136,13)
--cycle;
\path [draw=gray144, fill=c0, semithick]
(axis cs:1.60266666666667,10)
--(axis cs:1.864,10)
--(axis cs:1.864,13)
--(axis cs:1.60266666666667,13)
--(axis cs:1.60266666666667,10)
--cycle;
\path [draw=gray144, fill=c5, semithick]
(axis cs:1.86933333333333,11)
--(axis cs:2.13066666666667,11)
--(axis cs:2.13066666666667,15)
--(axis cs:1.86933333333333,15)
--(axis cs:1.86933333333333,11)
--cycle;
\path [draw=gray144, fill=c10, semithick]
(axis cs:2.136,14)
--(axis cs:2.39733333333333,14)
--(axis cs:2.39733333333333,18)
--(axis cs:2.136,18)
--(axis cs:2.136,14)
--cycle;
\path [draw=gray144, fill=c0, semithick]
(axis cs:2.60266666666667,10)
--(axis cs:2.864,10)
--(axis cs:2.864,13)
--(axis cs:2.60266666666667,13)
--(axis cs:2.60266666666667,10)
--cycle;
\path [draw=gray144, fill=c5, semithick]
(axis cs:2.86933333333333,13)
--(axis cs:3.13066666666667,13)
--(axis cs:3.13066666666667,18)
--(axis cs:2.86933333333333,18)
--(axis cs:2.86933333333333,13)
--cycle;
\path [draw=gray144, fill=c10, semithick]
(axis cs:3.136,17)
--(axis cs:3.39733333333333,17)
--(axis cs:3.39733333333333,21)
--(axis cs:3.136,21)
--(axis cs:3.136,17)
--cycle;
\draw[draw=gray144,fill=c0,line width=0.3pt] (axis cs:0,0) rectangle (axis cs:0,0);
\addlegendimage{ybar,ybar legend,draw=c0,fill=c0,line width=0.3pt}
\addlegendentry{$c=0$}

\draw[draw=gray144,fill=c5,line width=0.3pt] (axis cs:0,0) rectangle (axis cs:0,0);
\addlegendimage{ybar,ybar legend,draw=c5,fill=c5,line width=0.3pt}
\addlegendentry{$c=5$}

\draw[draw=gray144,fill=c10,line width=0.3pt] (axis cs:0,0) rectangle (axis cs:0,0);
\addlegendimage{ybar,ybar legend,draw=c10,fill=c10,line width=0.3pt}
\addlegendentry{$c=10$}

\addplot [draw=chartreuse1312550, fill=chartreuse1312550, mark=*, only marks]
table{%
x  y
};
\addplot [draw=aqua0255245, fill=aqua0255245, mark=*, only marks]
table{%
x  y
};
\addplot [draw=darkviolet1130255, fill=darkviolet1130255, mark=*, only marks]
table{%
x  y
};
\addplot [semithick, gray144, forget plot]
table {%
-0.266666666666667 11
-0.266666666666667 8
};
\addplot [semithick, gray144, forget plot]
table {%
-0.266666666666667 13
-0.266666666666667 16
};
\addplot [semithick, gray144, forget plot]
table {%
-0.332 8
-0.201333333333333 8
};
\addplot [semithick, gray144, forget plot]
table {%
-0.332 16
-0.201333333333333 16
};
\addplot [semithick, gray144, forget plot]
table {%
0 8
0 6
};
\addplot [semithick, gray144, forget plot]
table {%
0 11
0 15
};
\addplot [semithick, gray144, forget plot]
table {%
-0.0653333333333333 6
0.0653333333333333 6
};
\addplot [semithick, gray144, forget plot]
table {%
-0.0653333333333333 15
0.0653333333333333 15
};
\addplot [semithick, gray144, forget plot]
table {%
0.266666666666667 11
0.266666666666667 8
};
\addplot [semithick, gray144, forget plot]
table {%
0.266666666666667 14
0.266666666666667 18
};
\addplot [semithick, gray144, forget plot]
table {%
0.201333333333333 8
0.332 8
};
\addplot [semithick, gray144, forget plot]
table {%
0.201333333333333 18
0.332 18
};
\addplot [semithick, gray144, forget plot]
table {%
0.733333333333333 10
0.733333333333333 8
};
\addplot [semithick, gray144, forget plot]
table {%
0.733333333333333 13
0.733333333333333 17
};
\addplot [semithick, gray144, forget plot]
table {%
0.668 8
0.798666666666667 8
};
\addplot [semithick, gray144, forget plot]
table {%
0.668 17
0.798666666666667 17
};
\addplot [semithick, gray144, forget plot]
table {%
1 9
1 6
};
\addplot [semithick, gray144, forget plot]
table {%
1 12
1 16
};
\addplot [semithick, gray144, forget plot]
table {%
0.934666666666667 6
1.06533333333333 6
};
\addplot [semithick, gray144, forget plot]
table {%
0.934666666666667 16
1.06533333333333 16
};
\addplot [semithick, gray144, forget plot]
table {%
1.26666666666667 13
1.26666666666667 9
};
\addplot [semithick, gray144, forget plot]
table {%
1.26666666666667 16
1.26666666666667 20
};
\addplot [semithick, gray144, forget plot]
table {%
1.20133333333333 9
1.332 9
};
\addplot [semithick, gray144, forget plot]
table {%
1.20133333333333 20
1.332 20
};
\addplot [semithick, gray144, forget plot]
table {%
1.73333333333333 10
1.73333333333333 8
};
\addplot [semithick, gray144, forget plot]
table {%
1.73333333333333 13
1.73333333333333 17
};
\addplot [semithick, gray144, forget plot]
table {%
1.668 8
1.79866666666667 8
};
\addplot [semithick, gray144, forget plot]
table {%
1.668 17
1.79866666666667 17
};
\addplot [semithick, gray144, forget plot]
table {%
2 11
2 7
};
\addplot [semithick, gray144, forget plot]
table {%
2 15
2 21
};
\addplot [semithick, gray144, forget plot]
table {%
1.93466666666667 7
2.06533333333333 7
};
\addplot [semithick, gray144, forget plot]
table {%
1.93466666666667 21
2.06533333333333 21
};
\addplot [semithick, gray144, forget plot]
table {%
2.26666666666667 14
2.26666666666667 10
};
\addplot [semithick, gray144, forget plot]
table {%
2.26666666666667 18
2.26666666666667 24
};
\addplot [semithick, gray144, forget plot]
table {%
2.20133333333333 10
2.332 10
};
\addplot [semithick, gray144, forget plot]
table {%
2.20133333333333 24
2.332 24
};
\addplot [semithick, gray144, forget plot]
table {%
2.73333333333333 10
2.73333333333333 8
};
\addplot [semithick, gray144, forget plot]
table {%
2.73333333333333 13
2.73333333333333 17
};
\addplot [semithick, gray144, forget plot]
table {%
2.668 8
2.79866666666667 8
};
\addplot [semithick, gray144, forget plot]
table {%
2.668 17
2.79866666666667 17
};
\addplot [semithick, gray144, forget plot]
table {%
3 13
3 8
};
\addplot [semithick, gray144, forget plot]
table {%
3 18
3 25
};
\addplot [semithick, gray144, forget plot]
table {%
2.93466666666667 8
3.06533333333333 8
};
\addplot [semithick, gray144, forget plot]
table {%
2.93466666666667 25
3.06533333333333 25
};
\addplot [semithick, gray144, forget plot]
table {%
3.26666666666667 17
3.26666666666667 11
};
\addplot [semithick, gray144, forget plot]
table {%
3.26666666666667 21
3.26666666666667 27
};
\addplot [semithick, gray144, forget plot]
table {%
3.20133333333333 11
3.332 11
};
\addplot [semithick, gray144, forget plot]
table {%
3.20133333333333 27
3.332 27
};
\addplot [semithick, gray144, forget plot]
table {%
-0.397333333333333 13
-0.136 13
};
\addplot [semithick, gray144, forget plot]
table {%
-0.130666666666667 9
0.130666666666667 9
};
\addplot [semithick, gray144, forget plot]
table {%
0.136 12
0.397333333333333 12
};
\addplot [semithick, gray144, forget plot]
table {%
0.602666666666667 12
0.864 12
};
\addplot [semithick, gray144, forget plot]
table {%
0.869333333333333 11
1.13066666666667 11
};
\addplot [semithick, gray144, forget plot]
table {%
1.136 14
1.39733333333333 14
};
\addplot [semithick, gray144, forget plot]
table {%
1.60266666666667 12
1.864 12
};
\addplot [semithick, gray144, forget plot]
table {%
1.86933333333333 13
2.13066666666667 13
};
\addplot [semithick, gray144, forget plot]
table {%
2.136 16
2.39733333333333 16
};
\addplot [semithick, gray144, forget plot]
table {%
2.60266666666667 12
2.864 12
};
\addplot [semithick, gray144, forget plot]
table {%
2.86933333333333 16
3.13066666666667 16
};
\addplot [semithick, gray144, forget plot]
table {%
3.136 19
3.39733333333333 19
};
\end{axis}

\end{tikzpicture}
\vspace{-30pt}
  \caption{Total compression and decompression time vs. the compression configuration (in terms of parameters $c$ and $q$) for~Draco.}
  \label{fig:draco_compression_params}
\end{figure}

\subsubsection{(De)compression time}
In Fig.~\ref{fig:gpcc_compression_params} we show the
total compression and decompression time for G-PCC as a
function of the compression configuration.
The median time increases as compression becomes more conservative, from around $150.4$ ms at p0 to $580.2$ ms at p3, i.e., proportionally with the PQS parameter and so with the number of points that remains in the point cloud. In fact, higher compression permits to discard more points via coarse quantization, which not only accelerates the encoding process, but also reduces the time required for the point cloud to be reconstructed during decoding. 
The top panel of Fig.~\ref{fig:codec_time_size} confirms that the (de)compression time grows linearly with the file size. 
Notice that, even in the most aggressive configuration (p0), the (de)compression time is always higher than $100$ ms which, as reported in Sec.~\ref{sec:related}, is the maximum tolerable \ac{e2e} delay for \ac{TD} applications according to the 3GPP specifications~\cite{3gpp.22.186}. 

\definecolor{retloss}{RGB}{31,84,140}
\begin{table*}[t]
\caption{3D AP (IoU $0.50$) on the extended SELMA testing dataset for the three available classes (cars, pedestrians, cyclists). Values in parentheses indicate the AP as a percentage of the corresponding uncompressed point cloud. 
Cell shading is also proportional to the AP degradation with respect to \ the uncompressed point cloud (darker $=$ higher degradation). Draco entries are averaged over the three compression levels $c$.}
\label{tab:retention}
\centering
\begin{adjustbox}{max width=\textwidth}
\footnotesize
\begin{tabular}{llccccccccccc}
\toprule
& & & \multicolumn{3}{c}{Car} & \multicolumn{3}{c}{Pedestrian} & \multicolumn{3}{c}{Cyclist} \\
\cmidrule(lr){4-6}\cmidrule(lr){7-9}\cmidrule(lr){10-12}
Codec & Cfg. & Size (KB) & PV-RCNN & SECOND & PointPillars & PV-RCNN & SECOND & PointPillars & PV-RCNN & SECOND & PointPillars \\
\midrule
-- & Uncompressed & -- & $93.5$ & $90.6$ & $88.9$ & $60.2$ & $56.6$ & $60.1$ & $26.2$ & $31.9$ & $22.5$ \\
\midrule
G-PCC & p0 & $1.9$ & \cellcolor{retloss!3}$89.8$ $(96\%)$ & $87.6$ $(97\%)$ & \cellcolor{retloss!17}$65.2$ $(73\%)$ & \cellcolor{retloss!38}$24.7$ $(41\%)$ & \cellcolor{retloss!35}$25.7$ $(45\%)$ & \cellcolor{retloss!52}$12.2$ $(20\%)$ & \cellcolor{retloss!8}$23.1$ $(88\%)$ & \cellcolor{retloss!6}$29.0$ $(91\%)$ & \cellcolor{retloss!45}$6.8$ $(30\%)$ \\
G-PCC & p1 & $5.0$ & $91.8$ $(98\%)$ & $89.3$ $(98\%)$ & \cellcolor{retloss!6}$81.3$ $(91\%)$ & \cellcolor{retloss!13}$48.5$ $(81\%)$ & \cellcolor{retloss!11}$46.9$ $(83\%)$ & \cellcolor{retloss!24}$37.8$ $(63\%)$ & \cellcolor{retloss!3}$24.8$ $(95\%)$ & $31.4$ $(98\%)$ & \cellcolor{retloss!10}$18.9$ $(84\%)$ \\
G-PCC & p2 & $15.9$ & $93.3$ $(100\%)$ & $89.3$ $(98\%)$ & \cellcolor{retloss!4}$83.9$ $(94\%)$ & \cellcolor{retloss!8}$52.7$ $(88\%)$ & \cellcolor{retloss!6}$51.6$ $(91\%)$ & \cellcolor{retloss!10}$51.0$ $(85\%)$ & \cellcolor{retloss!4}$24.7$ $(94\%)$ & $31.1$ $(97\%)$ & \cellcolor{retloss!5}$20.8$ $(92\%)$ \\
G-PCC & p3 & $33.4$ & $93.5$ $(100\%)$ & $90.5$ $(100\%)$ & $86.6$ $(97\%)$ & \cellcolor{retloss!3}$57.7$ $(96\%)$ & $55.4$ $(98\%)$ & $59.2$ $(98\%)$ & $26.0$ $(99\%)$ & $32.0$ $(100\%)$ & $22.4$ $(100\%)$ \\
\midrule
Draco & 8xx & $9.4$ & \cellcolor{retloss!28}$53.7$ $(57\%)$ & \cellcolor{retloss!37}$39.0$ $(43\%)$ & \cellcolor{retloss!47}$24.7$ $(28\%)$ & \cellcolor{retloss!65}$0.2$ $(0\%)$ & \cellcolor{retloss!64}$0.5$ $(1\%)$ & \cellcolor{retloss!65}$0.1$ $(0\%)$ & \cellcolor{retloss!54}$4.6$ $(18\%)$ & \cellcolor{retloss!53}$5.9$ $(18\%)$ & \cellcolor{retloss!59}$2.0$ $(9\%)$ \\
Draco & 9xx & $15.4$ & \cellcolor{retloss!5}$85.9$ $(92\%)$ & \cellcolor{retloss!7}$80.6$ $(89\%)$ & \cellcolor{retloss!16}$66.7$ $(75\%)$ & \cellcolor{retloss!48}$15.6$ $(26\%)$ & \cellcolor{retloss!47}$15.6$ $(28\%)$ & \cellcolor{retloss!58}$6.2$ $(10\%)$ & \cellcolor{retloss!12}$21.2$ $(81\%)$ & \cellcolor{retloss!13}$25.8$ $(81\%)$ & \cellcolor{retloss!37}$9.6$ $(43\%)$ \\
Draco & 10xx & $24.6$ & $92.0$ $(98\%)$ & $89.1$ $(98\%)$ & \cellcolor{retloss!4}$83.9$ $(94\%)$ & \cellcolor{retloss!11}$50.1$ $(83\%)$ & \cellcolor{retloss!8}$49.4$ $(87\%)$ & \cellcolor{retloss!17}$44.1$ $(73\%)$ & \cellcolor{retloss!4}$24.7$ $(95\%)$ & $31.2$ $(98\%)$ & \cellcolor{retloss!11}$18.8$ $(84\%)$ \\
Draco & 11xx & $37.6$ & $93.4$ $(100\%)$ & $90.5$ $(100\%)$ & $86.6$ $(97\%)$ & \cellcolor{retloss!4}$56.9$ $(95\%)$ & \cellcolor{retloss!3}$53.9$ $(95\%)$ & $58.7$ $(98\%)$ & $25.3$ $(97\%)$ & $31.7$ $(99\%)$ & $22.6$ $(100\%)$ \\
\bottomrule
\end{tabular}
\end{adjustbox}
\end{table*}

Similarly, in Fig.~\ref{fig:draco_compression_params} we show the total compression and decompression time for Draco as a function of $q$ and $c$. Specifically, this time increases with $q$. This is due to the fact that the resulting point cloud is represented with more bits, which requires more time for both encoding and decoding. Similarly the (de)compression time also increases with the compression level $c$, especially when  $q$ is small. This is because a higher $c$ enables additional predictive and entropy-coding stages, which reduce the size of the encoded data at the cost of more computation.
As $q$ and $c$ increase, the (de)compression time also increases. Conversely, the size of the compressed point cloud decreases with $c$ but increases with $q$. Because of this antagonistic behavior, there is no explicit correlation between the (de)compression time and the size of the resulting point cloud, as illustrated in the bottom panel of Fig.~\ref{fig:codec_time_size}. Moreover, we can see that the decompression time is almost constant when the file size is more than $\SI{20}{KB}$ (it varies from $\SI{5.4}{ms}$ to $\SI{6.6}{ms}$), while the compression time does not have a clear trend.

Importantly, the (de)compression time with Draco is generally orders of magnitude lower than with G-PCC. In the best case, Draco can compress data in less than $10$ ms, against more than $150$ ms for the best G-PCC configuration. Considering that LiDAR sensors generally capture data at $30$ fps, i.e., one perception every around $33$ ms, Draco is capable of processing data in real-time, that is within the frame rate of the LiDAR.
Finally, from Fig.~\ref{fig:codec_time_size} we also observe that decoding is consistently faster than encoding for both codecs; this is a positive feature for TD since processing tasks such as object detection can be executed sooner after data is received and~decompressed.


\subsection{3D Object Detection Results}
\label{sub:detection-performance}

In Table~\ref{tab:retention} we evaluate the performance of the object detectors described in Sec.~\ref{sub:detectors} in terms of the \ac{AP} metric for each class available in SELMA, namely cars, pedestrians, and cyclists.
We use a threshold of 0.50 for the \ac{IoU}, meaning that only bounding boxes with an IoU greater
than 0.50 are considered for the AP computation. 
For G-PCC, we consider all the four options, i.e., p0, p1, p2, p3. For Draco, we only evaluate the impact of the number of quantization bits $q\in\{8,9,10,11\}$, while averaging over all values of~$c$, given the minor impact of $c$ with respect to $q$, as discussed in Sec.~\ref{sub:compression-results}. We use the notation ``$q$xx'' to indicate that the performance of Draco depends only on $q$.

First, we observe that the \ac{AP} increases when lower compression is applied, for any detector and codec. This is expected as a lighter compression preserves more structural information and details in the point cloud, so detectors can identify objects in the scene more accurately. However, the AP does not increase indefinitely, and eventually reaches a plateau. At this point, the error introduced by compression is negligible. This is a desirable feature for \ac{TD} applications, as it ensures that the decompressed point cloud remains virtually identical to the original training data.

Next, we note that G-PCC outperforms Draco in terms of \ac{AP}, while also reducing the compressed file size as illustrated in Fig.~\ref{fig:comparison_compressors}. This is because G-PCC uses geometric methods based on Octrees and voxels to compress data, which preserve the spatial structure of the point cloud more effectively. 
Moreover, we observe that PointPillars is the weakest detector on cars and cyclists, since it collapses 3D point clouds into pseudo-images using pillars, which lose depth information and deteriorate object detection. 
In contrast, PV-RCNN and SECOND achieve comparable detection results, although SECOND is more effective on more compressed point clouds, i.e., when the task is more challenging. This can be due to sparse convolutions in SECOND, which can be effective even in extreme cases.

\definecolor{retloss}{RGB}{31,84,140}
\begin{table*}[t]
\caption{Total average \ac{e2e} latency for receiving and processing \ac{LiDAR} frames, together with the corresponding compression and decompression, inference, and transmission times. Cells are shaded and marked with \ding{55} when the total latency is higher than 100 ms, i.e., the 3GPP requirement for \ac{TD} applications~\cite{3gpp.22.186}. Draco entries are averaged over the three compression levels $c$. The notation n/a indicates that transmission is not feasible on the \ac{V2X}~network.}
\label{tab:budget}
\centering
\begin{adjustbox}{max width=\textwidth}
\footnotesize
\begin{tabular}{llcccccccccc}
\toprule
& & & & & \multicolumn{2}{c}{PV-RCNN} & \multicolumn{2}{c}{SECOND} & \multicolumn{2}{c}{PointPillars} \\
\cmidrule(lr){6-7}\cmidrule(lr){8-9}\cmidrule(lr){10-11}
Codec & Cfg. & Size (KB) & \makecell{Compr. + \\ decompr. (ms)} & Tx (ms) & Infer. (ms) & Total (ms) & Infer. (ms) & Total (ms) & Infer. (ms) & Total (ms) \\
\midrule
-- & Uncompressed & $\approx 1066$ & -- & n/a & $168.2$ & \cellcolor{retloss!28}n/a \ding{55} & $17.4$ & \cellcolor{retloss!28} n/a \ding{55} & $42.3$ & \cellcolor{retloss!28} n/a \ding{55} \\
\midrule
G-PCC & p0 & $1.9$ & $106.4$ & $7.3$ & $91.9$ & \cellcolor{retloss!28}$205.7$ \ding{55} & $19.0$ & \cellcolor{retloss!28}$132.8$ \ding{55} & $40.8$ & \cellcolor{retloss!28}$154.5$ \ding{55} \\
G-PCC & p1 & $5.0$ & $160.8$ & $9.4$ & $99.8$ & \cellcolor{retloss!28}$270.0$ \ding{55} & $19.0$ & \cellcolor{retloss!28}$189.2$ \ding{55} & $40.7$ & \cellcolor{retloss!28}$210.9$ \ding{55} \\
G-PCC & p2 & $15.9$ & $349.8$ & $15.4$ & $105.0$ & \cellcolor{retloss!28}$470.2$ \ding{55} & $19.9$ & \cellcolor{retloss!28}$385.1$ \ding{55} & $41.2$ & \cellcolor{retloss!28}$406.4$ \ding{55} \\
G-PCC & p3 & $33.4$ & $621.2$ & $20.6$ & $123.7$ & \cellcolor{retloss!28}$765.6$ \ding{55} & $20.5$ & \cellcolor{retloss!28}$662.3$ \ding{55} & $41.6$ & \cellcolor{retloss!28}$683.4$ \ding{55} \\
\midrule
Draco & 8xx & $9.4$ & $11.4$ & $12.2$ & $169.3$ & \cellcolor{retloss!28}$192.8$ \ding{55} & $19.7$ & $43.3$ \ding{51} & $40.1$ & $63.7$ \ding{51} \\
Draco & 9xx & $15.4$ & $12.4$ & $15.1$ & $169.3$ & \cellcolor{retloss!28}$196.8$ \ding{55} & $19.7$ & $47.2$ \ding{51} & $40.1$ & $67.6$ \ding{51} \\
Draco & 10xx & $24.6$ & $13.8$ & $18.3$ & $169.3$ & \cellcolor{retloss!28}$201.4$ \ding{55} & $19.7$ & $51.8$ \ding{51} & $40.1$ & $72.2$ \ding{51} \\
Draco & 11xx & $37.6$ & $15.5$ & $21.5$ & $169.3$ & \cellcolor{retloss!28}$206.2$ \ding{55} & $19.7$ & $56.7$ \ding{51} & $40.1$ & $77.1$ \ding{51} \\
\bottomrule
\end{tabular}
\end{adjustbox}
\end{table*}

Finally, Table~\ref{tab:retention} also shows that vulnerable road users such as pedestrians and cyclists are considerably harder to detect than cars.
For example, using the same threshold of $0.50$ for the IoU and PV-RCNN, the AP on uncompressed data for the pedestrians (cyclists) class is $60.2$ ($26.2$) against $93.5$ for the car class, a gap of roughly $33$ ($66$) AP points.
This effect is due to the fact that pedestrians and cyclists are smaller and may be represented by fewer points, so they are more difficult to detect. At the
same time, these classes are less common than cars in the
SELMA dataset (class imbalance problem~\cite{oksuz2019imbalance}), which further
deteriorates the AP performance.

\section{Experimental Results for V2X Teleoperated Driving Performance}
\label{sec:network-results}
According to the 3GPP specifications for TD applications~\cite{3gpp.22.186}, sensor data must be received and processed within an \ac{e2e} latency of up to $100$ ms. This delay comprises multiple components, including data compression and decompression, inference for object detection, and data transmission. The latter depends on the underlying \ac{V2X} network.
 In Sec.~\ref{sec:results}, we only evaluated the trade-off between point cloud compression efficiency and object detection accuracy. In this section, we extend the analysis to evaluate the \ac{e2e} latency of the complete \ac{TD} pipeline, and provide
guidelines for selecting the most suitable compression-detection configurations.

\subsection{Simulation Parameters}
We simulate data transmission using the \ac{V2X} protocol stack in ns-3, one of the most popular and accurate discrete-event simulators~\cite{riley2010ns}.
 We consider an urban vehicular scenario with $5$ \acp{UE} acting as connected vehicles transmitting high-resolution \ac{LiDAR} frames at 30 fps to a remote teleoperator. The size of frames is modeled based on SELMA, considering different codecs.
 Mobility traces are generated using \ac{SUMO} \cite{sumo}, mapped onto real urban layouts extracted from \ac{OSM}.
    The wireless channel between the UEs and the V2X gNB, co-located with the teleoperator, is modeled using \ac{GEMV2} \cite{gemv}. Specifically, channel traces are pre-calculated to capture realistic large-scale fading and shadowing effects, relying on empirical channel datasets collected in the city of Bologna, in Italy, as described in~\cite{drago2022artificialintelligencevehicularwireless,11087616}. 
    We use NR V2X numerology 3 for transmissions at \ac{FR2} at $28\text{ GHz}$,   
    with a system bandwidth of $50\text{ MHz}$ and a transmission power of $23\text{ dBm}$. At the transport layer, we use the User Datagram Protocol (UDP) to minimize protocol overhead, which is critical for real-time control.
    We run simulations of 80 s, and compute full-stack performance metrics such as the \ac{e2e} throughput and latency.

    To simulate data compression, decompression, and inference, we use the compressors and detectors described in Sec.~\ref{sec:background}, trained as described in Sec.~\ref{sec:training_details}.

\subsection{Latency Results}
Table~\ref{tab:budget} reports the total \ac{e2e} latency required to receive and process sensors frames, together with the individual contributions of compression and decompression time, transmission time, and inference time for object detection. We consider 24 different compression-detection configurations.
We observe that transmitting the uncompressed point cloud (of size 1066 KB) at 30 fps generates an average data rate of around $250$ Mbps/UE, which exceeds the capacity of the simulated link. Therefore, compression is a precondition for data transmission, which motivates our research analysis.

\subsubsection{Compression and decompression time}
Table~\ref{tab:budget} shows that the sum of the compression and decompression time using G-PCC ranges between $106.4$ ms and $621.2$ ms based on the level of compression. Draco, instead, can compress data in less than 15 ms for all compression configurations.
These results confirm our findings in Fig.~\ref{fig:gpcc_compression_params}.

\subsubsection{Inference time}
The inference time depends on several factors, including the
detector architecture (depth, width, and number
of parameters) and size, and the type of hardware (CPU, GPU, TPU
accelerator).  
Table~\ref{tab:budget} shows that PV-RCNN is the slowest detector, though the inference time can be reduced using G-PCC by up to approximately 46\%
compared to Draco. In fact, PV-RCNN discards points that
are in the same position after quantization, thus reducing the
number of points to be processed by the detector.
On the contrary, the performance of PointPillars and SECOND is almost independent of the codec, since both detectors operate on grids of fixed dimension; the inference time is around 40 and 20 ms, respectively.

\subsubsection{Transmission time}
Table~\ref{tab:budget} shows that the transmission time on the V2X network using G-PCC is up to 40\% lower than Draco, given that the former produces smaller compressed point
clouds, with a minimum size of 1.9 KB for p0 (vs. 0.4 KB for 8xx).

\begin{figure}[t!]
  \centering
  \definecolor{gpccblue}{HTML}{2A78D6}
\definecolor{dracoorange}{HTML}{EB6834}

\begin{tikzpicture}
\begin{semilogxaxis}[
    width=\columnwidth,
    height=0.70\columnwidth,
    xlabel={End-to-end per-frame latency [ms]},
    ylabel={Pedestrian AP (IoU 0.50)},
    xmin=35, xmax=900,
    ymin=-3, ymax=66,
    xtick={50,100,200,400,800},
    xticklabels={50,100,200,400,800},
    log ticks with fixed point,
    ytick={0,10,20,30,40,50,60},
    grid=both,
    grid style={line width=0.2pt, draw=gray!25},
    tick label style={font=\footnotesize},
    label style={font=\footnotesize},
    legend columns=2,
    legend cell align=left,
    legend style={font=\scriptsize, at={(0.5,1.02)}, anchor=south,
                  draw=none, fill=none, column sep=4pt, row sep=-1pt},
]
 
\fill[black!6] (axis cs:35,-3) rectangle (axis cs:100,66);
\draw[dashed, gray!70, thick] (axis cs:100,-3) -- (axis cs:100,66);
\node[font=\scriptsize, anchor=north east, text=gray!70]
      at (axis cs:97,65) {3GPP TD};
 
\addplot[color=gpccblue, mark=*, mark size=1.8pt, thin]
  table[x=lat, y=ped, row sep=\\] {
  size  rate   lat    car   ped   cyc  \\
  1.9   0.056  205.7  89.8  24.7  23.1 \\
  5.0   0.146  270.0  91.8  48.5  24.8 \\
  15.9  0.466  470.2  93.3  52.7  24.7 \\
  33.4  0.979  765.6  93.5  57.7  26.0 \\
  };
\addlegendentry{PV-RCNN, G-PCC}

\addplot[color=dracoorange, mark=*, mark size=1.8pt, thin]
  table[x=lat, y=ped, row sep=\\] {
  size  rate   lat    car   ped   cyc  \\
  9.4   0.275  192.8  53.7   0.2   4.6 \\
  15.4  0.451  196.8  85.9  15.6  21.2 \\
  24.6  0.721  201.4  92.0  50.1  24.7 \\
  37.6  1.102  206.2  93.4  56.9  25.3 \\
  };
\addlegendentry{PV-RCNN, Draco}

\addplot[color=gpccblue, mark=triangle*, mark size=2.2pt, thin]
  table[x=lat, y=ped, row sep=\\] {
  size  rate   lat    car   ped   cyc  \\
  1.9   0.056  132.8  87.6  25.7  29.0 \\
  5.0   0.146  189.2  89.3  46.9  31.4 \\
  15.9  0.466  385.1  89.3  51.6  31.1 \\
  33.4  0.979  662.3  90.5  55.4  32.0 \\
  };
\addlegendentry{SECOND, G-PCC}

\addplot[color=dracoorange, mark=triangle*, mark size=2.2pt, thin]
  table[x=lat, y=ped, row sep=\\] {
  size  rate   lat    car   ped   cyc  \\
  9.4   0.275   43.3  39.0   0.5   5.9 \\
  15.4  0.451   47.2  80.6  15.6  25.8 \\
  24.6  0.721   51.8  89.1  49.4  31.2 \\
  37.6  1.102   56.7  90.5  53.9  31.7 \\
  };
\addlegendentry{SECOND, Draco}
 
\addplot[color=gpccblue, mark=square*, mark size=1.7pt, thin]
  table[x=lat, y=ped, row sep=\\] {
  size  rate   lat    car   ped   cyc  \\
  1.9   0.056  154.5  65.2  12.2   6.8 \\
  5.0   0.146  210.9  81.3  37.8  18.9 \\
  15.9  0.466  406.4  83.9  51.0  20.8 \\
  33.4  0.979  683.4  86.6  59.2  22.4 \\
  };
\addlegendentry{PointPillars, G-PCC}

\addplot[color=dracoorange, mark=square*, mark size=1.7pt, thin]
  table[x=lat, y=ped, row sep=\\] {
  size  rate   lat    car   ped   cyc  \\
  9.4   0.275   63.7  24.7   0.1   2.0 \\
  15.4  0.451   67.6  66.7   6.2   9.6 \\
  24.6  0.721   72.2  83.9  44.1  18.8 \\
  37.6  1.102   77.1  86.6  58.7  22.6 \\
  };
\addlegendentry{PointPillars, Draco}
 
\end{semilogxaxis}
\end{tikzpicture}
\vspace{-20pt}
  \caption{Pedestrian AP (IoU 0.50) vs. the total \ac{e2e} latency of
    Table~\ref{tab:budget}, for each codec and detector. Markers
    along a curve correspond, left to right, to increasingly conservative
    compression (p0--p3 for G-PCC, 8xx--11xx for Draco). The gray area corresponds to an \ac{e2e} latency lower than 100 ms, i.e., the 3GPP requirement for \ac{TD} applications~\cite{3gpp.22.186}.}
    \label{fig:feasibility}
\end{figure}

\subsection{Compression-Detection Configuration Guidelines}\label{ssec:operating}
In Fig.~\ref{fig:feasibility} we combine the total \ac{e2e} latency reported in Table~\ref{tab:budget} with the AP for pedestrians, i.e., one of the most complex detection classes in SELMA, for each codec and detector. Markers
    along a curve correspond, left to right, to increasingly conservative
    compression (p0--p3 for G-PCC, 8xx--11xx for Draco).
    We observe that only eight compression-detection combinations guarantee a total \ac{e2e} latency below 100 ms, i.e., the 3GPP requirement for \ac{TD} applications~\cite{3gpp.22.186}, namely all four Draco compressors paired with either SECOND ($43.3$-$56.7$ ms) or PointPillars ($63.7$-$77.1$ ms). 
    Of these, Draco 8xx and 9xx provide an AP of only 0.5 and 15.6, respectively, making these configurations unsuitable in the TD scenario.
    Four combinations remain, namely Draco 10xx and 11xx with SECOND or PointPillars.
    The choice between them should be based on the class of objects to detect: while SECOND is faster than PointPillars (up to $56.7$ vs. $77.1$ ms), it achieves higher accuracy for cyclists ($\text{AP}=31.7$ vs. $22.6$), while PointPillars performs better for pedestrians ($\text{AP}=58.7$ vs. $53.9$). Performance on cars is comparable ($\text{AP}=90.5$ vs. $86.6$).
From these results, we conclude that Draco is the best
approach to satisfy \ac{TD} requirements, when the main constraint is defined in terms of the maximum tolerable \ac{e2e} latency.

\begin{table}[t]
\centering
\caption{Recommended codec and detector as a function of the~constraints.}
\label{tab:guidelines}
\footnotesize
\setlength{\tabcolsep}{4pt}
\begin{adjustbox}{max width=0.45\textwidth}
\begin{tabular}{@{}llccc@{}}
\toprule
Binding constraint & Best codec + detector & Delay & Rate & Ped./Cyc. \\
 & & [ms] & [MBps] & AP \\
\midrule
Latency, cyclists first    & Draco 11xx + SECOND       & $56.7$  & $1.10$ & $53.9$ / $31.7$ \\
Latency, pedestrians first & Draco 11xx + PointPillars & $77.1$  & $1.10$ & $58.7$ / $22.6$ \\
Latency and capacity           & Draco 10xx + SECOND       & $51.8$  & $0.72$ & $49.4$ / $31.2$ \\
Capacity                     & G-PCC p2 + SECOND         & $385.1$ & $0.47$ & $51.6$ / $31.1$ \\
Capacity (severe)             & G-PCC p1 + SECOND         & $189.2$ & $0.15$ & $46.9$ / $31.4$ \\
\bottomrule
\end{tabular}
\end{adjustbox}
\end{table}

However, the price of feasibility is bandwidth. 
The data rate at 30 fps required by Draco 10xx and 11xx is $0.72$ MBps and $1.10$ MBps per UE, respectively, against only $0.15$ MBps for G-PCC p1, with a comparable detection accuracy for pedestrian ($\text{AP}=46.9)$.
As such, G-PCC remains preferable in capacity-constrained networks with limited transmission resources, where processing delays can be tolerated.
In this case, the relatively high compression and decompression delays of G-PCC can be mitigated by operating at the hardware level, for example using an optimized Octree codec instead of the reference TMC13 implementation~\cite{wang2024octree}. 
Moreover, our inference time is measured with an unoptimized PyTorch implementation, whereas pillar- and voxel-based detectors deployed on NVIDIA TensorRT have been shown to provide an additional speedup by at least a factor of two~\cite{wang2023dsvt,zhou2023fastpillars}. 

Table~\ref{tab:guidelines} reports the recommended codecs and detectors as a function of the critical constraints to satisfy, i.e., \ac{e2e} delay, capacity, and/or the target object class to prioritize during~detection.

\section{Conclusion}
\label{sec:conclusions}
In this paper we conducted an extensive simulation campaign to evaluate the performance of different compression and object detection algorithms (namely Draco and G-PCC, and PV-RCNN, PointPillars and SECOND, respectively), and the resulting impact on the V2X network considering a \ac{TD} application. We trained and tested these algorithms on the open-source SELMA dataset, that we modified to incorporate bounding boxes for the car, pedestrian, and cyclist classes, considering LiDAR point clouds. We evaluated the compression efficiency, measured in terms of the resulting file size after compression, the (de)compression time, the inference time, the detection quality, measured in terms of AP, and the \ac{e2e} transmission delay and data rate. We draw several~conclusions:
\begin{itemize}
    \item G-PCC is the most effective codec in terms of compression rate and accuracy, reducing the point cloud to only $1.9$ KB at p0, with an average higher AP than Draco at every file size. 
    However, the (de)compression time of its underlying Octree architecture in the reference software ranges from $106.4$ ms to $621.2$ ms, so no configuration we tested satisfies the 3GPP \ac{e2e} delay requirement for TD applications once inference is accounted for.    
    \item Draco compresses and decompresses in only $11$-$16$ ms, but only two configurations (10xx and 11xx) preserve good detection accuracy. Combined with SECOND or PointPillars, the total \ac{e2e} delay required to transmit and process the point cloud ranges between $51.8$ and $77.1$ ms. 
    \item In the feasibility regions, the choice of the detector depends on the class of objects to detect: PointPillars is stronger on pedestrians ($\text{AP}=58.7$ vs. $53.9$) while SECOND on cyclists ($\text{AP}=31.7$ against $22.6$). 
    PV-RCNN requires at least $168.2$ ms for inference even on uncompressed point clouds, and is not usable for \ac{TD}.
    \item Overall, Draco is the preferred approach when the main constraint is defined in terms of the e2e delay, while G-PCC is desirable for capacity-constrained
networks as it permits to reduce the size of the data to
transit, and uses less channel resources.
\end{itemize}

As part of our future work, we will extend the evaluation to the low-latency branches of G-PCC, namely L3C2 and predictive geometry, and to real-time learned codecs.
Moreover, we will repeat the latency measurements using TensorRT, so as to separate the performance of the coders from that of their reference implementations. We will also design and implement advanced solutions to automatically determine the optimal compression and detection configuration based on the underlying network performance and TD requirements.

\bibliographystyle{IEEEtran}
\bibliography{bibs}

\end{document}